%% file: main.tex
\documentclass{article} %
\usepackage{iclr2027_conference,times}

\input{math_commands.tex}

\usepackage{float}       %
\usepackage{silence}     %
\usepackage{hyperref}
\usepackage{url}
\usepackage{graphicx}
\usepackage{booktabs}
\usepackage{tikz}
\usetikzlibrary{shapes.geometric}   %
\usepackage{algorithm}
\usepackage{algpseudocode}   %
\renewcommand{\floatpagefraction}{0.8}
\let\origtable\table
\renewcommand{\table}{\setlength{\belowcaptionskip}{6pt}\origtable}

\newcommand{\lgdentry}[4]{%
  \tikz[baseline=-0.5ex]{\draw[#1,line width=#2,#3](0,0)--(1.5em,0);}\,{#4}}

\newcommand{\lgdmark}[4]{%
  \tikz[baseline=-0.6ex]{\node[#2,draw=#1,fill=#3,inner sep=0pt,minimum size=4.5pt,line width=0.4pt]{};}\,{#4}}

\title{Values Shape What Self-Improving Coding Agents Build}

\author{Jason Liang \quad Dylan Waldner \quad Risto Miikkulainen \\
Cognizant AI Lab \\
\texttt{\{jason.liang,dylan.waldner,risto\}@cognizant.com}
}

\iclrfinalcopy
\begin{document}

\maketitle
\fancyhead{}

\begin{abstract}
Self-improving agents are typically optimized for the results they produce, with less attention to how they produce them. Their instructions give designers limited ways to shape how they work, especially when the task is open-ended. This paper studies whether a short value statement in an agent's persistent instruction file changes not only how well the agent performs but also what it builds. Four higher-order value types in Schwartz's theory of basic human values were evaluated: \emph{openness to change}, \emph{conservation}, \emph{self-enhancement}, and \emph{self-transcendence}. Across diverse coding tasks, the values were held fixed within each run over multiple iterations of self-improvement. Three findings emerged: (1) Values matter for performance across a diverse set of tasks. The \emph{openness} values increase performance on held-out data, whereas the opposite \emph{conservation} values result in scores below those of an agent given no values. (2) Values change what agents build, producing solution algorithms that differ in size, complexity and approach. (3) Human oversight is still needed for designing values. Allowing the agent to revise its own values narrows performance differences between them, and values distilled from the AI literature provide no detectable gain over having no values. Properly designed values can thus influence what a self-improving coding agent builds and, where the task permits, how well it performs.
\end{abstract}

\section{Introduction}
\label{sec:introduction}

Systems that improve their own code have a long history in AI research \citep{schmidhuber2003godel, zelikman2024stop}. This paper studies their most recent form: a large language model (LLM) agent that iteratively improves a solver, i.e.\ a program that produces a solution for a task, and the tools around it. The agent works in sessions, each starting from the code the last one committed, so that improvements accumulate through repeated coding and evaluation. Self-improvement matters because it lifts the ceiling of a single attempt: gains can compound from session to session without a person in the loop. In this manner, LLMs editing programs inside an evolutionary search have discovered algorithms that beat handwritten ones \citep{romeraparedes2024funsearch, novikov2025alphaevolve} and improved machine-found solutions at a lower sample cost \citep{lange2026shinkaevolve}. Agents that rewrite their own implementations go further and improve themselves \citep{zhang2025dgm, robeyns2025sica}. As these systems run longer with less supervision, what they produce may drift from what their designers intended, or it can stay boxed in by the instructions they were first given. Therefore, how they improve matters as much as how far they improve: the means matter, not only the ends.

Existing methods for improving agents and prompts use task performance to evaluate progress, with behavior steered by fixed, manually written meta-agent prompt templates \citep{hu2025adas}, prompts the agent can edit itself \citep{robeyns2025sica}, or instructions optimized for task accuracy \citep{yang2024opro}. Yet within each session the agent makes many choices that the score does not specify: which part of the solver to rewrite, whether to replace an algorithm or only tune it, and when to keep or discard a change. The score reflects what counts as improvement but says little about how to achieve it, so these choices fall to whatever dispositions the model brings. A value statement is task-general guidance for the decisions that the task and its evaluation metric leave open. It is a natural instrument for shaping those dispositions, and it can have a large effect on the final outcome.

Whether a value statement changes what an agent builds has not been tested. Prior designs do not isolate the contribution of values from the task specification, the method guidance and the score-based selection, and the literature surveyed in Section~\ref{sec:related-work} contains no controlled experiment that isolates it. Therefore, the question this paper asks is whether different values alter what the agent builds under a shared self-improvement protocol. At stake is how much of the result is decided by a disposition it is explicitly given, as opposed to the task itself and the model's own biases.

As shown by Figure~\ref{fig:schwartz}, one useful source of agent dispositions is Schwartz's theory of basic human values \citep{schwartz1992universals, schwartz2012overview}, a widely used and empirically grounded map of what people value. It suits the purpose for two reasons. It defines values as goals that transcend any one situation, so the same values can serve any task. Its ten basic values group into four higher-order types on two opposing axes, \emph{openness} against \emph{conservation} and \emph{self-enhancement} against \emph{self-transcendence}. In this manner, each type has an opposite to test against. This paper explores all four types by inserting short, carefully designed value statements into the agent's instruction file at every iteration of an agentic self-improvement loop. The four Schwartz values are compared against two controls, no values and a placebo of similar length, across six diverse coding tasks: ALE-Bench, Circle Packing, Entity Matching, Game 2048, Symbolic Regression, and SZ-Tetris. In Circle Packing, the best generated solutions exceed AlphaEvolve and ShinkaEvolve at $n=26$ \citep{novikov2025alphaevolve,lange2026shinkaevolve} and beat records across many other sizes (Tables~\ref{tab:compare-packing} and~\ref{tab:packing-records}).

\begin{figure}[!t]
\centering
\begin{tikzpicture}
\node[anchor=south west, inner sep=0] (img) at (0,0) {\includegraphics[width=\linewidth]{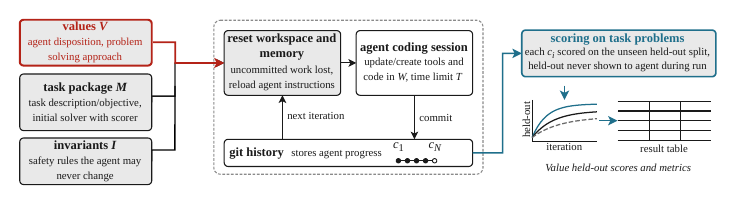}};
\begin{scope}[x={(img.south east)}, y={(img.north west)}]
\node[anchor=north, inner sep=0, font=\footnotesize] at (0.140,0.062) {(\textit{a}) Pre-run agent instructions};
\node[anchor=north, inner sep=0, font=\footnotesize] at (0.468,0.062) {(\textit{b}) Self-improvement loop};
\node[anchor=north, inner sep=0, font=\footnotesize] at (0.842,0.062) {(\textit{c}) Post-run evaluation and scoring};
\end{scope}
\end{tikzpicture}\vspace{2pt}
\caption{\textbf{Value-based self-improvement framework.} (\textit{a}) The framework fixes the values, task package and safety invariants. (\textit{b}) Each iteration restores workspace $W$ to the preceding commit, reloads the instructions, runs a coding session and saves its code. (\textit{c}) A separate scorer evaluates saved solvers on held-out data outside $W$ and tracks how fixed values shape code and performance across iterations.}
\label{fig:pipeline}
\end{figure}

The main findings of this paper are as follows:
\begin{enumerate}\itemsep0pt
\item \textbf{Values matter for performance across a diverse set of tasks.} Relative to the no-values control, values can shift scores in both directions. Furthermore, the opposing values of \emph{openness} and \emph{conservation} perform significantly differently from one another.
\item \textbf{Values affect the approach used for solving each task.} Under different values the agent writes solvers that differ in algorithms and heuristics on the same task. Across tasks, held-out performance tracks solver size and complexity more closely than evaluation count, API cost, or total run time.
\item \textbf{Human oversight is still needed for designing values.} Allowing the agent to revise its own values narrows performance differences between them: the best-performing values decline while the worst-performing improve. Values distilled from the AI literature provide no detectable gain over having no values.
\end{enumerate}

The paper is organized as follows. Section~\ref{sec:related-work} reviews related work in LLM-based coding, self-improvement, and values. Section~\ref{sec:methods} describes the methodology in detail, and Section~\ref{sec:results} presents the experimental results within and across tasks. Section~\ref{sec:discussion} evaluates the three contributions and proposes future work on value-based self-improvement. The appendices provide protocol details, value specifications, task details with published comparisons, cross-task statistics, solver algorithm analyses, and empirical evaluations of several framework variations.

\section{Related Work}
\label{sec:related-work}

This section summarizes related literature on three fronts: (1) program evolution, (2) self-improving agents, and (3) values and personas in language models. These research areas form the basis for the self-improvement framework described later.

\textbf{LLM-driven program evolution.} This research area uses LLMs as the mutation or crossover operator of an evolutionary loop over programs \citep{lehman2022elm, meyerson2024lmx, hemberg2024evolving, shojaee2025llmsr}. FunSearch found new cap-set constructions and online bin-packing heuristics \citep{romeraparedes2024funsearch}, while AlphaEvolve discovered improvements for problems such as Circle Packing \citep{novikov2025alphaevolve}. ShinkaEvolve achieved excellent results in Circle Packing at a lower sample cost and improved existing solutions for ALE-Bench \citep{lange2026shinkaevolve, imajuku2025alebench}. A parallel research direction investigates prompt optimization against a fixed task objective \citep{fernando2024promptbreeder, yang2024opro}. Together these systems show that LLM-driven search is effective on difficult optimization problems.

\textbf{Self-improving and self-modifying agents.} The G\"odel machine formalized a problem solver that rewrites its own code once a proof shows the rewrite leads to improvement \citep{schmidhuber2003godel}. STOP improves the scaffold that calls the model \citep{zelikman2024stop}, and ADAS searches agent designs written as code \citep{hu2025adas}. The G\"odel Agent, the Darwin G\"odel Machine and SICA let an agent rewrite its own logic, prompts, or harness while guided by performance on a task \citep{yin2025godelagent, zhang2025dgm, robeyns2025sica}. Hyperagents combine the task agent and the meta agent that edits it in one program, so the procedure that produces the next self-modification is itself editable \citep{zhang2026hyperagents}. In this paper, the harness and the instructions are fixed in order to determine the impact of fixed values on the performance and behavior of a self-improving agent.

\textbf{Values, personas and prompt sensitivity.} The four values follow the higher-order types of Schwartz's theory of basic human values \citep{schwartz1992universals, schwartz2012overview}. In people, the values-behavior link is strong for some values and marginal for others \citep{bardi2003values}. Constitutional AI put a written list of principles into training \citep{bai2022constitutional}, whereas here a fixed model reads a written value statement at inference. Role-play prompts help on some reasoning tasks \citep{kong2024roleplay}, while system-prompt personas did not improve average accuracy on factual questions \citep{zheng2024helpfulassistant}. Prior work has measured personality-like responses, opinion skew and value profiles in LLMs, including profiles based on Schwartz's values \citep{serapiogarcia2023personality, jiang2023inducing, santurkar2023opinions, rozen2025consistent, ren2024valuebench, yao2024fulcra, huang2025wild}. A written value statement acts on existing model behavior, and this paper measures its effects.

\section{Methodology}
\label{sec:methods}

This section describes the self-improvement framework, illustrated in Figure~\ref{fig:pipeline}, with pseudocode in Algorithm~\ref{alg:loop}. It consists of three stages, described in the subsections below: (1) Before each run, the framework sets the agent's instructions, including the values under test. (2) During the run, the agent works in one workspace for multiple iterations and only committed work survives. (3) After the run, a separate scorer scores each iteration's saved solver on data kept outside the workspace.

\subsection{Agent Instructions Setup}
\label{sec:instructions}

\textbf{The instruction file.} The file the agent reads at the start of every session is composed of the following sections: the values, the invariants (safety rules that must be followed) and a pointer to the task description. The instruction file uses a common structure across runs. Furthermore, to prevent hacking or jailbreaking, the loop restores the values and the invariants every iteration to their initial condition. Appendix~\ref{app:protocol} shows a concrete example of the instruction file.

\textbf{Task description and invariants.} The task description gives the objective, a scorer for the development (dev) split, a starting solver and a baseline reference score. The invariants require the agent to verify its work, keep it inside the workspace, and leave the benchmark and scoring unchanged. Scripted checks test the task files for answer leakage and scoring integrity.

\begin{figure}[!t]
\centering
\begin{tikzpicture}
\node[anchor=south west, inner sep=0] (img) at (0,0) {\includegraphics[width=\linewidth]{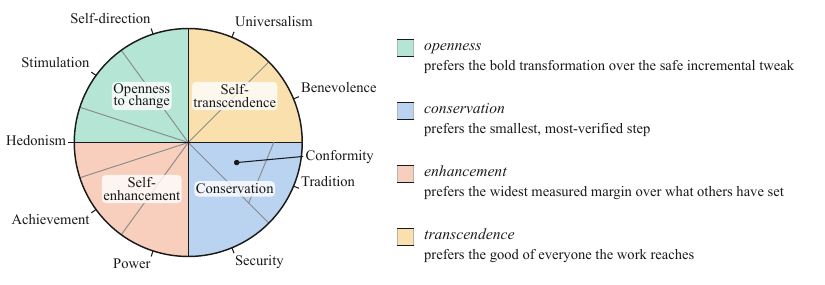}};
\begin{scope}[x={(img.south east)}, y={(img.north west)}]
\node[anchor=north, inner sep=0, font=\footnotesize] at (0.227,0.040) {(\textit{a})};
\node[anchor=north, inner sep=0, font=\footnotesize] at (0.721,0.040) {(\textit{b})};
\end{scope}
\end{tikzpicture}\vspace{2pt}
\caption{\textbf{Schwartz's values and their use as agent instructions.} (\textit{a}) A visualization of the ten basic human values and four higher-order types \citep{schwartz2012overview}. (\textit{b}) Summaries of the four value statements given to the agents (Appendix~\ref{app:values}). The experiments compare two opposing pairs: \emph{openness} versus \emph{conservation}, and \emph{enhancement} versus \emph{transcendence}.}
\label{fig:schwartz}
\end{figure}

\textbf{The values.} Each of the four values (Figure~\ref{fig:schwartz}) represents one of Schwartz's higher-order types and serves as guidance for choosing the next change to the solver: \emph{openness}, for instance, asks the agent to ``prefer the bold transformation \ldots{} over the safe incremental tweak''. In the experiments, these values are specified as a list of short bullet points that have similar overall lengths. Their wording was drafted with an LLM from Schwartz's accounts of the higher-order types \citep{schwartz1992universals, schwartz2012overview} and edited by the authors over several rounds of careful manual review. Appendix~\ref{app:values} lists these value statements in detail.

\textbf{The controls.} There are two controls for comparing the values against. The \emph{none} control states in one sentence that the agent has no values. Because the question is whether having values changes the outcome, this control is the reference for every comparison in Table~\ref{tab:results} and answers whether having values influences what agents build and code. The \emph{placebo} control uses the same structure and a similar length but is composed of off-task text unrelated to the values. This second control checks whether the length of the values text alone could cause changes in agent behavior (Table~\ref{tab:contrasts}).

\subsection{The Self-Improvement Loop}
\label{sec:loop}

\textbf{Tracking code changes with Git.} The loop is driven by the Claude Code agent harness \citep{anthropic2025claudecode}, using \texttt{claude-opus-5} at high reasoning effort with no recorded web search or fetch calls (Table~\ref{tab:protocol}). Each iteration resets the workspace to the code and notes from the last commit and reloads the instruction file with values and invariants from the start of the run (Algorithm~\ref{alg:loop}). The agent runs under a preset time limit that, once reached, forcibly terminates the coding session. The agent must follow the protocol the instruction file sets out: read the log of earlier iterations, choose the one change that most advances the values, implement and verify it against the provided scorer, log the results and finally commit the change. The agent may edit the solver, write tools, and run the dev scorer, but only what it commits survives. It carries no conversational state or memory between sessions.

\textbf{Everything else stays fixed.} Only the solver code, tools, and documentation change between iterations, while the framework, the model, the task package and the fixed instruction sections stay unchanged for the run. The framework provides dev scores during the run and held-out scores afterward. The agent is never told that other sets of values exist or that a comparison is under way: the instruction file purposely keeps the values and task isolated.

\subsection{Task Evaluation and Scoring}
\label{sec:evaluation}

\textbf{Evaluation metrics.} Each task has a dev split the agent can see and a held-out split kept outside its workspace (Appendix~\ref{app:tasks}). After the run, a separate scorer scores each iteration's saved solver on the held-out split and writes a per-iteration held-out curve. The primary statistic reported is the held-out score from the final iteration, averaged over multiple runs of the same value. Another statistic is the value's mean area under the held-out curve (AUC).

The framework also logs the number of scorer evaluations, API cost, and total elapsed run time, as well as various code complexity metrics of the solver being improved by the agent (Appendix~\ref{app:code-metrics}).

\textbf{Effect size and permutation tests.} To compare runs under different values, the paper reports an effect size and a $p$ value. The effect size is Cliff's delta $\delta$ \citep{cliff1993dominance}, the chance that values outscore the control, minus the reverse. Because $\delta$ is rank-based, it depends on score ordering rather than magnitude. The $p$ value comes from a two-sided permutation test \citep{davison1997bootstrap} on the difference in means: the share of 10,000 relabelings whose absolute difference is at least the observed one. This test assumes exchangeable labels under the null, without requiring a parametric score distribution. For comparisons pooled over all tasks, scores are z-score normalized within each task to put them on a common scale (Appendix~\ref{app:cross-task}). Because $\delta$ compares ranks and the permutation test compares means, a large $\delta$ does not necessarily imply a statistically significant $p$, as indicated by several results in Table~\ref{tab:results}.

\begin{figure}[!t]
\centering
\includegraphics[width=\linewidth]{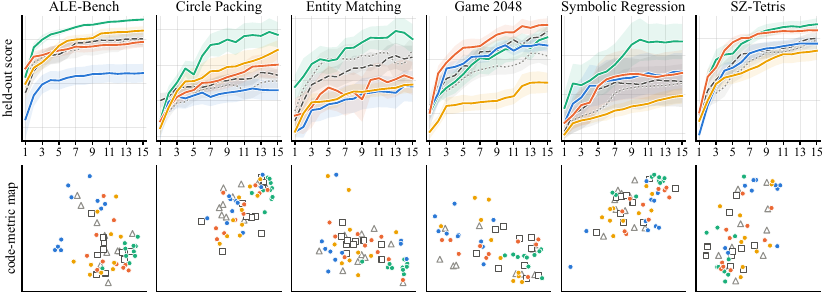}\\[2pt]
{\footnotesize\input{figures/legends/legend_values.tex}}
\caption{\textbf{Performance curves and solver code across six tasks.} Top: Mean held-out scores across iterations, with standard error bands and separate scales for each task. Bottom: t-SNE maps of final solvers based on code metrics (Appendix~\ref{app:code-metrics}). Colors indicate values, and hollow markers denote controls. Agents given different values reach different scores and build solvers with different code structures. The size and pattern of these differences vary across tasks.}
\label{fig:results}
\end{figure}

\begin{table}[!t]
\centering\footnotesize\renewcommand{\arraystretch}{0.90}\setlength{\tabcolsep}{3.4pt}
\caption{\textbf{Held-out performance relative to the no-values control.} Rows report Cliff's $\delta$ and permutation $p$ for final scores and area under the curve (AUC). The all-tasks rows pool scores standardized within each task. Bold separately marks large effects ($|\delta|>0.43$) and statistically significant differences ($p<0.05$). In the pooled results, \emph{openness} scores above \emph{none}, while \emph{conservation} and \emph{transcendence} score below it.}
\label{tab:results}
\input{figures/tables/table_master_t.tex}
\end{table}

\section{Experimental Results}
\label{sec:results}

This section gives experimental results for the four Schwartz values and the two controls on six diverse coding tasks. Results are also compared across tasks and under several variations of the framework. Each set of values is evaluated across 10 independent runs of 15 iterations lasting up to 30 minutes each. Figure~\ref{fig:results} plots the held-out curves of the six tasks as well as the t-SNE maps of their final solvers \citep{vandermaaten2008tsne}. Table~\ref{tab:results} compares each set of values and \emph{placebo} to the \emph{none} control for each task.

\subsection{ALE-Bench}
\label{sec:ale}

\textbf{Task description.} ALE-Bench is a benchmark of heuristic optimization problems from programming contests \citep{imajuku2025alebench}. The specific task being tested is AHC039, a popular problem also evaluated by other LLM-guided solvers such as ShinkaEvolve \citep{lange2026shinkaevolve}. The problem requires the solver to draw one axis-aligned polygon that meets certain constraints \citep{atcoder2024ahc039}. The agent must output solution code that creates the polygon under a hard 2\,s per-seed CPU budget. The held-out score is the sum of the individual scores for the 150 official private test cases. More details about the task are shown in Appendix~\ref{app:ale}.

\textbf{Result.} The \emph{openness} values score best, significantly above the control, while the \emph{conservation} values perform worse (Table~\ref{tab:results}). This task shows the strongest separation of values with a wide margin between \emph{openness} and \emph{conservation} (Table~\ref{tab:vs-openness}). Against the control, the other sets of values show weaker effects. Furthermore, \emph{openness} exceeds both the ALE-Bench leaderboard entry for the same LLM and the first-place human-coded solution, sitting only slightly below the absolute best LLM results (Table~\ref{tab:compare-ale}).

\subsection{Circle Packing}
\label{sec:packing}

\textbf{Task description.} For each of the 37 odd sizes from 27 to 99, the agent must pack that many circles, of any radii, into the unit square so that the sum of the radii is as large as possible. At each size, the score is $-\log_{10}$ of the packing's relative gap to a reference record, clamped between zero and seven. The task score is the mean over all sizes, so progress is measured in the average number of \emph{digits} of gap closed. The agent improves a solver and commits one feasible packing per size. The held-out split comprises 38 \emph{even} packings and measures generalization to unseen sizes with a fixed budget of 60\,s for each (more information in Appendix~\ref{app:packing}).

\textbf{Result.} The \emph{openness} and \emph{transcendence} values do best and both score above controls, while \emph{conservation} does worst (Table~\ref{tab:results}). The \emph{openness} values also beat \emph{conservation} in head-to-head performance (Table~\ref{tab:contrasts}). The other sets of values show weaker effects in either direction. As shown in Tables~\ref{tab:compare-packing} and~\ref{tab:packing-records}, the packings discovered by the self-improvement framework exceed AlphaEvolve and ShinkaEvolve \citep{novikov2025alphaevolve, lange2026shinkaevolve} at $n=26$ while matching or beating many existing records on Packomania \citep{specht2026packomania}.

\begin{table}[!t]
\centering\footnotesize\renewcommand{\arraystretch}{0.90}\setlength{\tabcolsep}{3.4pt}
\caption{\textbf{Pairwise comparisons of opposing values and the two controls.} Rows compare final held-out scores using Cliff's $\delta$ and permutation $p$, with bold thresholds as in Table~\ref{tab:results}. In the pooled results, \emph{openness} versus \emph{conservation} shows the largest effect, followed by \emph{enhancement} versus \emph{transcendence}. The \emph{placebo} and \emph{none} controls show the smallest difference.}
\label{tab:contrasts}
\input{figures/tables/table_axes_compact.tex}
\end{table}

\subsection{Entity Matching}
\label{sec:matching}

\textbf{Task description.} Given two noisy product records, the solver estimates the probability that they refer to the same product. Performance is measured by the F1 score for identifying matching pairs. The dev and held-out splits for the task are from the WDC Products benchmark \citep{peeters2024wdc} using its 80\% hard ``corner-case'' setting, with all held-out entities unseen in the dev set. The agent constructs a solver that is trained on the labels from the dev split. The unseen entities make this task the cleanest generalization test of the six. More details about the task are reported in Appendix~\ref{app:matching}.

\textbf{Result.} The \emph{openness} values do best and beat \emph{conservation} head to head (Table~\ref{tab:vs-openness}). On the other hand, \emph{conservation} and \emph{transcendence} values have large negative effects relative to the control (Table~\ref{tab:results}). The other sets of values show weaker separation with the overall spread of scores between the values and controls being moderate. As shown in Table~\ref{tab:compare-matching}, the best discovered solutions from this paper are able to exceed existing results from the literature.

\subsection{Game 2048}
\label{sec:2048}

\textbf{Task description.} Game 2048 is a sliding-tile game on a 4 by 4 board in which tiles with equal numbers merge when pushed together. The score is the sum of merged tile values \citep{guei2022optimistic}. The agent writes a full-fledged solver (evaluator, search and time management) with a CPU budget of 1.0\,s per game. The held-out split is 30 fresh games using 30 seeds never seen by the agent during self-improvement. Appendix~\ref{app:2048} provides more details.

\textbf{Result.} The \emph{enhancement} values have the highest mean and a large advantage over \emph{transcendence} (Tables~\ref{tab:results} and~\ref{tab:contrasts}). The \emph{transcendence} and \emph{placebo} values score significantly below the control, while \emph{conservation} has a large negative effect. Published comparisons place the best generated solvers below the strongest trained networks (Table~\ref{tab:compare-2048}).

\begin{figure}[!t]
\centering
\includegraphics[width=\linewidth]{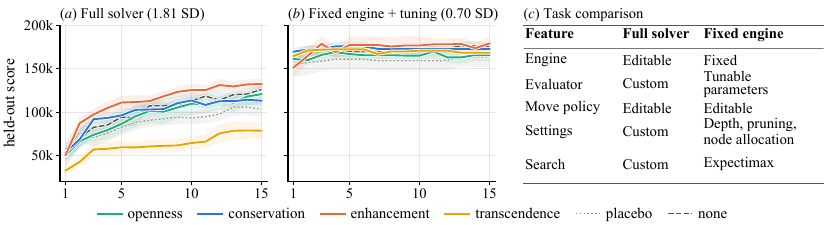}
\caption{\textbf{Game 2048 with a fully editable solver and a fixed engine.} Curves show mean held-out scores over iterations with standard error bands. Spread is the range of final means divided by the standard deviation (SD) of all final scores. The table summarizes differences between the two versions of Game 2048. Values show less separation with the fixed engine.}
\label{fig:2048-compare}
\end{figure}

\textbf{Fixed engine comparison.} A separate version uses an optimized expectimax engine adapted from \citet{xiao2014_2048ai}. The engine stays fixed while the agent tunes evaluator parameters, search depth, pruning and node allocation. Its mean scores change little across iterations and stay closer together than in the fully editable version (Figure~\ref{fig:2048-compare}).

\subsection{Symbolic Regression}
\label{sec:symreg}

\textbf{Task description.} Symbolic Regression seeks a closed-form expression that fits observed data. The agent tackles the 129 problems in the LSR-Synth snapshot used here \citep{shojaee2025llmsrbench,xia2026functionevolve}, which span a diverse range of scientific and mathematical formulas. The agent generates a program that proposes expressions through a metered evaluation call, which fits each expression's constants and returns its error, and the agent commits one expression per problem. The held-out evaluation refits each committed expression's constants and scores it on inputs outside the training range, testing range extrapolation of the discovered expression. More details about the task are in Appendix~\ref{app:symreg}.

\textbf{Result.} The \emph{openness} values rank first while \emph{transcendence} performs worst overall (Table~\ref{tab:results}). No set of values differs significantly from the control, although \emph{openness} has a large positive effect. A comparison between the best values and published results is shown in Table~\ref{tab:compare-symreg}, with the agent-discovered solutions performing around the middle of the pack.

\subsection{SZ-Tetris}
\label{sec:sz-tetris}

\textbf{Task description.} Stochastic SZ-Tetris is a specialized, harder variant of the popular block game \citep{szita2010sztetris, jaskowski2015sztetris, elfwing2018silu}. It is played on a 10 by 20 board with only S and Z pieces, drawn with equal probability and with no preview of the next piece allowed. A cleared row scores one point, and the game ends when a piece does not fit. The agent writes the whole controller under a budget of 1.0\,s of CPU per game. The held-out split is 100 fresh games from the same distribution on unseen seeds. Appendix~\ref{app:sz-tetris} contains more details about the task.

\textbf{Result.} The \emph{openness} values have the highest mean score (Table~\ref{tab:results}). \emph{Enhancement} beats \emph{transcendence} head to head and also outperforms the control by a smaller amount (Tables~\ref{tab:results} and~\ref{tab:contrasts}). The other sets of values show weaker effects. Published comparisons in Table~\ref{tab:compare-sz-tetris} again show trained neural solvers performing best while agent-discovered solvers score around average.

\subsection{Results Across Tasks}
\label{sec:cross-task}

\textbf{Performance differences between values.} Of the final-score comparisons between the control and other values in Table~\ref{tab:results}, one third show either a large effect size or statistical significance. Across all the tasks, on average \emph{openness} performs better than the control and \emph{conservation} performs worst. This contrast is further strengthened when a head-to-head comparison is done (Table~\ref{tab:contrasts}). The \emph{transcendence} values also tend to do worse than the control, although that varies by task, and \emph{enhancement} does not separate from the control all that much. There is a weaker but still noticeable separation between \emph{enhancement} and \emph{transcendence} in the head-to-head comparison.

\textbf{Different values built different solutions.} Analysis of the code for the solvers discovered on every task shows that \emph{openness} often creates larger, more complex solvers (Table~\ref{tab:cost-code}). Across multiple complexity measures, the solutions produced under \emph{openness} appear to exceed those produced under all other values on almost all the tasks. Similarly, \emph{conservation} tends to produce smaller, less complex solvers than the other values. Averaged across tasks, the held-out score tracks solver code size and complexity more closely than resource measures such as dev scorer evaluations, API cost or total run time (Table~\ref{tab:score-correlations}). As shown in Table~\ref{tab:code-neighbors}, a solver's nearest neighbors in code-metric space tend to come from its own set of values more often than chance predicts. Another interesting result is that solvers for different values start off close together but spread apart as the run iterations increase (Figure~\ref{fig:code-maps}).

\subsection{Alternative Values and Models}
\label{sec:follow-ups}

\textbf{Self-modification narrows differences between values.} On ALE-Bench, when the agent is allowed to modify and update its own values during every iteration, the spread across the values and controls shrinks roughly threefold (Figure~\ref{fig:self-mod}). The best and worst performing values, when allowed to self-modify, tend to converge to the middle: \emph{openness} falls and \emph{conservation} rises, each showing statistically significant differences from its non-self-modifying counterpart (Table~\ref{tab:self-mod}).

\textbf{Literature-derived values show no detectable gain.} Two new values are distilled from the AI and self-improvement literature, respectively. A deep research agent is assigned to review selected papers in AI and self-improvement research and then generate the values based on what it learned. On both the ALE-Bench and Game 2048 tasks, these derived values do not seem to have a large impact and both perform no better than the control (Appendix~\ref{app:literature}).

\textbf{Performance depends on value framing.} The \emph{openness} and \emph{conservation} value statements were each rewritten in three ethical framings: deontological (rules), utilitarian (outcomes), and virtue (character), retaining their original dispositions. All three \emph{openness} rewrites scored below the original, with statistically significant decreases for the deontological and utilitarian versions. The deontological \emph{conservation} rewrite had a large negative effect, although no \emph{conservation} rewrite differed significantly from the original (Appendix~\ref{app:framing}).

\textbf{Cheaper models perform noticeably worse.} The same two values were tested on two cheaper models, Claude Sonnet and Haiku. Both values scored significantly lower on these weaker models than on Opus, the main model used in this paper, with the lowest scores resulting from Haiku. The advantage of \emph{openness} over \emph{conservation} persisted on Sonnet but disappeared on Haiku with little improvement across iterations for either set of values (Appendix~\ref{app:models}).

\section{Discussion and Future Work}
\label{sec:discussion}

\textbf{Opposing sets of Schwartz values perform differently.} The clearest contrast is between \emph{openness} and \emph{conservation} on most tasks (Table~\ref{tab:contrasts}). This recurring advantage for \emph{openness} is consistent with a benefit from trying new approaches during solver revision. \emph{Enhancement} and \emph{transcendence} show weaker separation overall. The size of these differences varies across tasks.

\textbf{Open-ended solver design strengthens the effects of values.} In Game 2048, the spread of final mean scores across values is $1.81$ SD with a fully editable solver and $0.70$ SD with a fixed engine (Figure~\ref{fig:2048-compare}). The fixed engine version still allows hyperparameter and configuration tuning. This comparison suggests that freedom to choose or replace algorithms may give values more influence on performance.

\textbf{Values shape the solutions that agents build.} On every task, \emph{openness} produced longer, more complex solvers on average than \emph{none}, while \emph{conservation} produced shorter ones (Table~\ref{tab:cost-code}). A solver's nearest neighbors in code-metric space share its values well above chance level (Table~\ref{tab:code-neighbors}). Averaged across tasks, held-out scores correlate more closely with solver size and complexity than with evaluation count, API cost, or total run time (Table~\ref{tab:score-correlations}).

Appendix~\ref{app:deep-dive} examines the algorithms behind these differences. On ALE-Bench, every \emph{openness} solver uses free-cell or graph-cut search, while every \emph{conservation} solver retains column-based search. Circle Packing solvers mostly share a local optimizer but differ in how they generate and combine candidate packings. On both tasks, \emph{openness} runs favor exploration over exploitation and adopt more methods than \emph{conservation} runs. Their final solvers also receive lower mean maintainability ratings than those built under \emph{conservation}, although the ratings overlap.

\textbf{Human oversight is still needed for designing values.} Experiments with self-modifying and literature-derived values suggest that manual review and design remain important (Appendices~\ref{app:self-mod} and~\ref{app:literature}). A preliminary analysis finds that most self-modified values become highly specific to the task. The value statements also tend to grow to several times their original length. Literature-derived values remain concise, but are often cautious and tend toward \emph{conservation}. Deriving values from multiple papers may favor the safest, most common insights.

\textbf{Future work.} Value statements could draw from broader sources or become objects of search. An agent could be given the values of a school of philosophy, a professional code, or an influential thinker, extending the experiments in Appendix~\ref{app:follow-ups}. An outer evolutionary loop could also select and optimize values based on validation performance across multiple tasks. Jointly improving values and task solvers could yield value statements that transfer to unseen tasks and which people can inspect and reuse in their own agentic systems.

\section{Conclusion}
\label{sec:conclusion}

This paper investigated whether a value statement changes what a self-improving coding agent builds and how well it performs. The main findings are as follows: First, values can affect performance across diverse coding tasks: opposing values such as \emph{openness} and \emph{conservation} scored significantly differently from each other. Second, values shape what agents build, with certain values producing larger, more complex solvers across all six tasks. Third, human value design is still most effective: self-modification narrowed performance differences, while dispositions distilled from the literature offered no detectable gain. These results suggest that values can guide choices left open by the task objective, shaping both code and performance. As agents build on their previous work, choices encouraged by values become embedded in code that later iterations inherit. Properly designed values may therefore have lasting consequences on an agent's development, influencing which approaches it continues to explore and which alternatives it leaves behind.

\subsection*{Ethics statement}
\label{sec:ethics}

This study examines whether explicit values shape coding agents' behavior and performance. It uses public benchmarks and involves no human participants. The values are experimental instructions: higher benchmark scores do not establish that a value statement is ethically preferable or that an agent is safe or fair. Autonomous code changes require appropriate testing and human oversight before deployment. The experiments use a closed, paid model service: Anthropic's Claude API.

\subsection*{Reproducibility statement}
\label{sec:reproducibility}

Section~\ref{sec:methods} describes the self-improvement pipeline. Appendix~\ref{app:protocol} gives the main-experiment settings and fixed instruction sections, and Appendix~\ref{app:values} documents the value statements. Appendix~\ref{app:tasks} describes each task's scoring metrics, held-out split, reference scores and budget, while Appendix~\ref{app:cross-task} details the statistical comparisons. Appendices~\ref{app:solvers} and~\ref{app:deep-dive} describe the solver analyses, including the LLM review procedure, and Appendix~\ref{app:follow-ups} covers the follow-up experiments. Replication requires access to the specified proprietary models, and later model or API changes may affect the results.

\subsection*{AI use statement}
\label{sec:ai-use}

The experiments used Claude Code to generate and revise solver code (Section~\ref{sec:loop}). Generative AI tools, including Claude Code, assisted with drafting value statements, literature and citation review, analysis code, figure generation, interpretation of results, and drafting and revising the manuscript. LLM reviewers also classified solver algorithms and rated code maintainability (Appendix~\ref{app:deep-dive}). The authors reviewed and edited AI-assisted materials and take responsibility for the final text, claims, code and figures.

\input{main.bbl}
\input{appendix}
\end{document}

%% file: math_commands.tex
\usepackage{amsmath,amsfonts,bm}

\def\eqref#1{equation~\ref{#1}}

\def\1{\bm{1}}

\DeclareMathAlphabet{\mathsfit}{\encodingdefault}{\sfdefault}{m}{sl}
\SetMathAlphabet{\mathsfit}{bold}{\encodingdefault}{\sfdefault}{bx}{n}



%% file: figures/tables/table_master_t.tex
\begin{tabular}{@{}lrrrrrrrrrr@{}}
\toprule
 & \multicolumn{2}{c}{openness} & \multicolumn{2}{c}{conservation} & \multicolumn{2}{c}{enhancement} & \multicolumn{2}{c}{transcendence} & \multicolumn{2}{c}{placebo} \\
\cmidrule(lr){2-3}\cmidrule(lr){4-5}\cmidrule(lr){6-7}\cmidrule(lr){8-9}\cmidrule(lr){10-11}
Task & $\delta$ & $p$ & $\delta$ & $p$ & $\delta$ & $p$ & $\delta$ & $p$ & $\delta$ & $p$ \\
\midrule
\multicolumn{11}{@{}l}{\textit{Final score}} \\[1pt]
ALE-Bench & \textbf{+0.60} & \textbf{0.006} & \textbf{$-$0.64} & \textbf{0.005} & $-$0.06 & 0.711 & +0.30 & 0.277 & +0.12 & 0.957 \\
Circle Packing & \textbf{+0.60} & \textbf{0.010} & $-$0.24 & 0.214 & +0.12 & 0.470 & \textbf{+0.56} & \textbf{0.030} & $-$0.26 & 0.667 \\
Entity Matching & +0.28 & 0.361 & \textbf{$-$0.58} & 0.151 & $-$0.26 & 0.309 & \textbf{$-$0.64} & 0.074 & +0.26 & 0.654 \\
Game 2048 & $-$0.10 & 0.658 & \textbf{$-$0.46} & 0.101 & +0.20 & 0.435 & \textbf{$-$0.88} & \textbf{$<$0.001} & \textbf{$-$0.50} & \textbf{0.037} \\
Symbolic Regression & \textbf{+0.50} & 0.208 & +0.34 & 0.800 & +0.12 & 0.800 & +0.36 & 0.542 & +0.28 & 0.897 \\
SZ-Tetris & +0.38 & 0.092 & $-$0.14 & 0.534 & +0.30 & 0.204 & $-$0.16 & 0.273 & $-$0.16 & 0.822 \\
all tasks & +0.38 & \textbf{$<$0.001} & $-$0.27 & \textbf{0.003} & +0.08 & 0.653 & $-$0.16 & \textbf{0.043} & $-$0.06 & 0.355 \\
\midrule
\multicolumn{11}{@{}l}{\textit{AUC}} \\[1pt]
ALE-Bench & \textbf{+0.50} & \textbf{0.012} & \textbf{$-$0.56} & \textbf{0.011} & +0.10 & 0.917 & +0.26 & 0.431 & +0.12 & 0.976 \\
Circle Packing & \textbf{+0.56} & \textbf{0.022} & $-$0.20 & 0.292 & +0.04 & 0.940 & $-$0.08 & 0.488 & $-$0.24 & 0.330 \\
Entity Matching & \textbf{+0.46} & 0.110 & \textbf{$-$0.58} & 0.106 & $-$0.26 & 0.302 & \textbf{$-$0.62} & 0.064 & +0.16 & 0.543 \\
Game 2048 & $-$0.14 & 0.681 & $-$0.08 & 0.899 & \textbf{+0.48} & 0.133 & \textbf{$-$0.88} & \textbf{$<$0.001} & $-$0.20 & 0.234 \\
Symbolic Regression & \textbf{+0.56} & 0.288 & +0.24 & 0.830 & $-$0.02 & 0.843 & +0.26 & 0.465 & +0.32 & 0.807 \\
SZ-Tetris & +0.40 & 0.098 & $-$0.04 & 0.683 & +0.42 & 0.065 & $-$0.32 & 0.297 & $-$0.20 & 0.442 \\
all tasks & +0.40 & \textbf{$<$0.001} & $-$0.20 & \textbf{0.018} & +0.11 & 0.365 & $-$0.27 & \textbf{0.014} & $-$0.08 & 0.312 \\
\bottomrule
\end{tabular}

%% file: figures/tables/table_axes_compact.tex
\begin{tabular}{@{}lrrrrrr@{}}
\toprule
 & \multicolumn{2}{c}{openness vs conservation} & \multicolumn{2}{c}{enhancement vs transcendence} & \multicolumn{2}{c}{placebo vs none} \\
\cmidrule(lr){2-3}\cmidrule(lr){4-5}\cmidrule(lr){6-7}
Task & $\delta$ & $p$ & $\delta$ & $p$ & $\delta$ & $p$ \\
\midrule
ALE-Bench & \textbf{+1.00} & \textbf{$<$0.001} & $-$0.28 & 0.120 & +0.12 & 0.957 \\
Circle Packing & \textbf{+0.72} & \textbf{0.003} & $-$0.24 & 0.363 & $-$0.26 & 0.667 \\
Entity Matching & \textbf{+0.60} & \textbf{0.034} & +0.24 & 0.750 & +0.26 & 0.654 \\
Game 2048 & $-$0.02 & 0.545 & \textbf{+0.92} & \textbf{$<$0.001} & \textbf{$-$0.50} & \textbf{0.037} \\
Symbolic Regression & +0.30 & 0.254 & +0.00 & 0.396 & +0.28 & 0.897 \\
SZ-Tetris & \textbf{+0.54} & \textbf{0.025} & +0.38 & \textbf{0.046} & $-$0.16 & 0.822 \\
\midrule
all tasks & \textbf{+0.58} & \textbf{$<$0.001} & +0.24 & \textbf{0.024} & $-$0.06 & 0.355 \\
\bottomrule
\end{tabular}

%% file: appendix.tex
\clearpage
\appendix
\raggedbottom
\makeatletter\@addtoreset{table}{section}\@addtoreset{figure}{section}\makeatother
\renewcommand{\thetable}{\Alph{section}\arabic{table}}
\renewcommand{\thefigure}{\Alph{section}\arabic{figure}}
\renewcommand{\theHtable}{\thesection.\arabic{table}}
\renewcommand{\theHfigure}{\thesection.\arabic{figure}}
\setcounter{topnumber}{3}\setcounter{totalnumber}{5}
\renewcommand{\floatpagefraction}{0.8}
\makeatletter\setlength{\@fptop}{0pt}\setlength{\@fpsep}{8pt plus 4pt minus 8pt}\setlength{\@fpbot}{0pt plus 1fil}\makeatother   %

\newcommand{\slotboxsetup}{%
  \setlength{\fboxsep}{2pt}\setlength{\topsep}{1pt}\setlength{\partopsep}{0pt}%
  \setlength{\lineskip}{0pt}\setlength{\lineskiplimit}{0pt}%
  \widowpenalty=0 \clubpenalty=0 }

\section*{Appendix contents}
\noindent\begin{tabular}{@{}p{0.12\linewidth}p{0.84\linewidth}@{}}
\ref{app:protocol} & \textbf{Protocol details.} The self-improvement loop, instruction file, leakage checks, fixed prompt and settings used in the main experiments. \\
\ref{app:values} & \textbf{The values in detail.} The four Schwartz values, two controls and two literature-derived values. \\
\ref{app:tasks} & \textbf{Task details.} Held-out results and published comparisons for each task, with the setup, scoring budgets and qualifications needed to interpret them. \\
\ref{app:cross-task} & \textbf{Cross-task statistics.} Pooled score comparisons, pairwise tests between values and performance rankings across tasks. \\
\ref{app:solvers} & \textbf{What the agents built.} Solver size, resource use, code metrics and the best and worst solvers on each task. \\
\ref{app:deep-dive} & \textbf{Deep dive of discovered solvers.} Algorithm choices, exploration, maintainability ratings and method adoption on ALE-Bench and Circle Packing. \\
\ref{app:follow-ups} & \textbf{Follow-up experiments.} Self-modifying values, literature-derived values, ethical framing and model substitution. \\
\end{tabular}

\section{Protocol details}
\label{app:protocol}

Algorithm~\ref{alg:loop} specifies the three stages of Figure~\ref{fig:pipeline}: setup, coding iterations and held-out evaluation. Values and invariants stay fixed in the main experiments, with editable values tested separately in Appendix~\ref{app:self-mod}.

\input{figures/text/algorithm_loop.tex}

\textbf{Instructions and prompt.} The boxes below reproduce the fixed instruction sections for the main experiments. The values appear in Appendix~\ref{app:values}, and the omitted status block records the iteration, limits and run start time. The agent is instructed not to edit the fixed instruction sections. Each session receives the same prompt: read this file and the files it references, then complete exactly one iteration. The values guide which change the agent attempts at step 2 of the iteration protocol, subject to the fixed invariants and benchmark target.

\begingroup\slotboxsetup
\input{figures/text/instructions_file.tex}\par
\endgroup

\textbf{Shared settings.} Table~\ref{tab:protocol} lists the model and session settings. The starting solver and held-out split are fixed within each task. The generic instructions permit read-only web research. Search and fetch were available, with no calls recorded in the main runs.

\textbf{Leakage checks.} Three scripted guards scan the starting files for leaked answers. They check for held-out identifiers, names of winning methods, tuned parameter values, descriptions or code that could reconstruct the held-out split, and high-precision numbers from answer keys.

\textbf{Evaluation.} After each run, a separate scorer evaluates the saved solver from each iteration on held-out data. Final scores summarize the last iteration, while AUC summarizes performance across iterations. Results are reported across ten runs per set of values. Appendix~\ref{app:tasks} gives the task-specific scoring rules and CPU limits.

\begin{table}[!t]
\centering\footnotesize\setlength{\tabcolsep}{4pt}%
\caption{\textbf{Main configuration settings for the self-improvement loop.} The table lists the model, session limits and instruction settings for all six tasks. Each session inherits committed files but no conversational memory. While values are fixed, code improvements are allowed to accumulate.}
\label{tab:protocol}
\input{figures/tables/table_config_loop.tex}
\end{table}

\section{The values in detail}
\label{app:values}

This section shows the four Schwartz values, \emph{placebo} and \emph{none} from the main comparison, plus the two literature-derived values in Appendix~\ref{app:literature}. Each Schwartz value statement specifies default choices and priorities and explains how to approach the work. The placebo uses off-task text of similar length in the same three-bullet format. The \emph{none} control states in one sentence that the agent has no inherent values. Each box gives the word count, excluding bullets and standalone punctuation, with em dashes printed as colons.

\begingroup\slotboxsetup
\input{figures/text/values_texts.tex}\par
\endgroup

\section{Task details}
\label{app:tasks}

This section summarizes the setup and results for each task. The figures compare performance across values, and the tables place those results alongside published methods.

\subsection{ALE-Bench}
\label{app:ale}

\begin{figure}[H]
\centering
  \begin{minipage}[c]{0.49\linewidth}\centering\footnotesize\setlength{\tabcolsep}{3pt}%
    \input{figures/tables/table_heldout_ale-bench-ahc039.tex}  \end{minipage}\hfill
  \begin{minipage}[c]{0.49\linewidth}\centering
    \includegraphics[width=\linewidth]{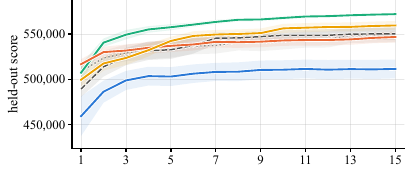}%
  \end{minipage}\\[2pt]
  {\footnotesize\input{figures/legends/legend_values.tex}}%
\caption{\textbf{Openness and conservation separate on ALE-Bench.} The table compares final scores with the no-values control (\emph{none}) using Cliff's $\delta$ and permutation $p$, and curves show held-out score over iterations with standard error bands. Bold separately marks $|\delta|>0.43$ and $p<0.05$. Openness scores significantly above \emph{none} and \emph{conservation} significantly below it, giving the clearest contrast between opposing values (Table~\ref{tab:contrasts}).}
\label{fig:ale}
\end{figure}

On AHC039, the agent improves a C++ solver that draws an axis-aligned polygon under a perimeter limit and a 2.0\,s CPU budget per case. The score counts enclosed mackerels minus enclosed sardines plus one, floored at zero. Figure~\ref{fig:ale} tests the resulting search procedure on the 150 official private cases and sums their scores.

Table~\ref{tab:compare-ale} compares scores with same-model leaderboard entries, other LLM systems and the winning human solution. The \emph{openness} mean exceeds both same-model entries and the winning human score, while the strongest listed LLM remains ahead.

\begin{table}[H]
\centering\footnotesize
\caption{\textbf{ALE-Bench scores approach the strongest published results.} The table compares AHC039 scores and API costs. Mean rows average scores and costs over ten runs per set of values. Best-run rows show the highest score for each value. Bold marks the highest score in each section. The \emph{openness} mean exceeds the same-model and winning human scores but trails the strongest listed LLM.}
\label{tab:compare-ale}
\input{figures/tables/table_sota_ale-bench-ahc039.tex}
\end{table}

\subsection{Circle Packing}
\label{app:packing}

\begin{figure}[H]
\centering
  \begin{minipage}[c]{0.49\linewidth}\centering\footnotesize\setlength{\tabcolsep}{3pt}%
    \input{figures/tables/table_heldout_circle-packing.tex}  \end{minipage}\hfill
  \begin{minipage}[c]{0.49\linewidth}\centering
    \includegraphics[width=\linewidth]{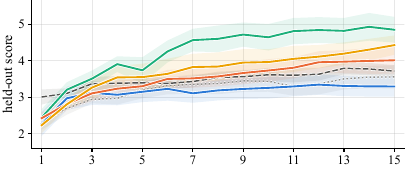}%
  \end{minipage}\\[2pt]
  {\footnotesize\input{figures/legends/legend_values.tex}}%
\caption{\textbf{Openness and transcendence lead on unseen packing sizes.} The table tests final scores against \emph{none}, and curves show held-out score with standard error bands (Figure~\ref{fig:ale}). Scores measure closeness to the reference on a logarithmic scale, averaged across held-out even sizes. Both leading values significantly beat the control, while \emph{conservation} ranks last.}
\label{fig:packing}
\end{figure}

The agent improves a Python solver that packs circles of arbitrary radii into the unit square to maximize their summed radii. It commits feasible packings for 37 visible odd sizes from 27 to 99. Figure~\ref{fig:packing} tests transfer to 38 unseen even sizes from 26 to 100, with 60\,s of CPU per size. Each size's score is $-\log_{10}$ of its relative gap to a fixed Packomania reference, clamped between zero and seven. The task score is their mean.

Table~\ref{tab:compare-packing} compares sums of radii at $n=26$ and $n=32$. The best \emph{openness} and \emph{none} packings match the reference records and exceed the listed AlphaEvolve and ShinkaEvolve constructions. Table~\ref{tab:packing-records} extends the comparison across sizes using a fixed Packomania snapshot \citep{specht2026packomania}. Openness matches or improves on the reference at more unseen sizes than \emph{none}, while the two have similar coverage on visible sizes.

\begin{table}[H]
\centering\footnotesize
\caption{\textbf{Solvers match Circle Packing records at held-out sizes.} The table compares sums of radii at $n=26$ and $n=32$, with mean and best-run scores from ten runs per set of values. Bold marks the highest score for each packing size. Published methods search each size separately, while the solvers in this paper were run on held-out sizes. The best \emph{openness} and \emph{none} runs match Packomania within tolerance and exceed the listed AlphaEvolve and ShinkaEvolve results.}
\label{tab:compare-packing}
\input{figures/tables/table_sota_circle-sizes.tex}
\end{table}

\begin{table}[H]
\centering\footnotesize
\caption{\textbf{Solvers reach Packomania records across many packing sizes.} Each cell lists two counts: sizes where a run ties the Packomania record and sizes where a run exceeds it \citep{specht2026packomania}. If any run exceeds a record, that size appears only in the second count. \emph{Openness} ties or exceeds records at more held-out sizes than \emph{none}, with similar counts on dev sizes.}
\label{tab:packing-records}
\input{figures/tables/table_sota_circle-records.tex}
\end{table}

\subsection{Entity Matching}
\label{app:matching}

\begin{figure}[H]
\centering
  \begin{minipage}[c]{0.49\linewidth}\centering\footnotesize\setlength{\tabcolsep}{3pt}%
    \input{figures/tables/table_heldout_entity-matching.tex}  \end{minipage}\hfill
  \begin{minipage}[c]{0.49\linewidth}\centering
    \includegraphics[width=\linewidth]{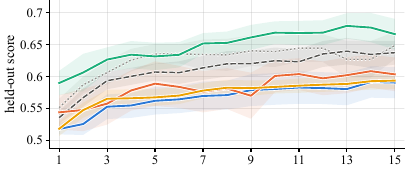}%
  \end{minipage}\\[2pt]
  {\footnotesize\input{figures/legends/legend_values.tex}}%
\caption{\textbf{Openness leads on unseen product entities.} The table tests final F1 against \emph{none}, and curves show held-out score with standard error bands (Figure~\ref{fig:ale}). Held-out entities are absent from dev. Openness beats \emph{conservation} head to head (Table~\ref{tab:vs-openness}), while \emph{conservation} and \emph{transcendence} score below the control with large effect sizes.}
\label{fig:matching}
\end{figure}

The agent improves a Python matcher that predicts whether two records refer to the same product, training it on visible pairs. The experiment uses WDC Products at its 80\% hard ``corner-case'' setting, with all held-out entities absent from dev \citep{peeters2024wdc}. Figure~\ref{fig:matching} measures positive-class F1 on the published gold standard of 4{,}500 pairs under a 120\,s CPU limit, testing transfer to new entities.

Table~\ref{tab:compare-matching} compares mean and best-run F1 with the hand-built Magellan matcher and the neural models RoBERTa, Ditto and HierGAT. The best \emph{openness} and \emph{none} runs exceed all listed references. Their means exceed Magellan but remain below the neural models, showing a substantial gap between typical and best-run performance.

\begin{table}[H]
\centering\footnotesize
\caption{\textbf{Entity Matching scores compared with published results.} Scores are F1 percentages on the same held-out test set, with mean and best-run scores from ten runs per set of values. Bold marks the highest score in each section. The published neural models were not restricted to the 120\,s CPU limit used here. The best \emph{openness} and \emph{none} runs exceed all listed published scores, but their means remain below the neural models' scores.}
\label{tab:compare-matching}
\input{figures/tables/table_sota_entity-matching.tex}
\end{table}

\subsection{Game 2048}
\label{app:2048}

\begin{figure}[H]
\centering
  \begin{minipage}[c]{0.49\linewidth}\centering\footnotesize\setlength{\tabcolsep}{3pt}%
    \input{figures/tables/table_heldout_game-2048-c-v2.tex}  \end{minipage}\hfill
  \begin{minipage}[c]{0.49\linewidth}\centering
    \includegraphics[width=\linewidth]{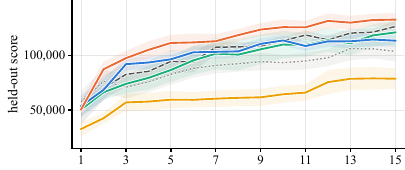}%
  \end{minipage}\\[2pt]
  {\footnotesize\input{figures/legends/legend_values.tex}}%
\caption{\textbf{Enhancement has the highest mean on Game 2048.} The table tests final scores against \emph{none}, and curves show held-out score with standard error bands (Figure~\ref{fig:ale}). Scores average the sum of merged tile values over fresh games under the per-game CPU cap. The strongest contrast is between \emph{enhancement} and \emph{transcendence} (Table~\ref{tab:contrasts}).}
\label{fig:2048}
\end{figure}

The agent writes the complete C solver for the standard 4 by 4 game, including search, evaluation and time management, under a 1.0\,s CPU budget per game. Scores sum merged tile values until game over or the solver budget is exhausted. Figure~\ref{fig:2048} reports the mean over 30 fresh games on unseen seeds from the dev spawn distribution.

Table~\ref{tab:compare-2048} compares game scores with an LLM-written solver and trained n-tuple networks. The \emph{enhancement} and \emph{none} means exceed the listed LLM-written solver's score and fall between those of the two smaller networks. The larger trained networks remain well ahead.

\begin{table}[H]
\centering\footnotesize
\caption{\textbf{Game 2048 scores fall between those of smaller trained networks.} The table compares mean and best-run scores from ten runs per set of values, measured by the sum of merged tile values. Bold marks the highest score in each section. Training budgets and CPU limits differ across published methods. The \emph{enhancement} and \emph{none} means exceed the listed LLM-written solver's score but remain far below those of the larger networks.}
\label{tab:compare-2048}
\input{figures/tables/table_sota_game-2048-c-v2.tex}
\end{table}

\subsection{Symbolic Regression}
\label{app:symreg}

\begin{figure}[H]
\centering
  \begin{minipage}[c]{0.49\linewidth}\centering\footnotesize\setlength{\tabcolsep}{3pt}%
    \input{figures/tables/table_heldout_symreg-lsr.tex}  \end{minipage}\hfill
  \begin{minipage}[c]{0.49\linewidth}\centering
    \includegraphics[width=\linewidth]{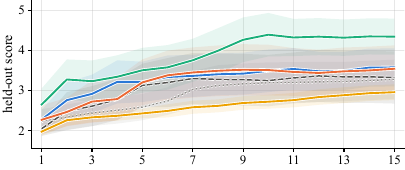}%
  \end{minipage}\\[2pt]
  {\footnotesize\input{figures/legends/legend_values.tex}}%
\caption{\textbf{Symbolic Regression shows limited separation between values.} The table tests final scores against \emph{none}, following Figure~\ref{fig:ale}. Curves show held-out score with standard error bands. Openness has the highest mean and \emph{transcendence} the lowest, with wide variation across runs.}
\label{fig:symreg}
\end{figure}

The agent improves a search program for the 129 LSR-Synth problems, submitting expressions to a metered call that fits their constants and returns their error \citep{shojaee2025llmsrbench,xia2026functionevolve}. It commits one expression per problem. Held-out scoring refits the constants and tests inputs outside the training range (Figure~\ref{fig:symreg}). The score averages $-\log_{10}(\mathrm{NMSE})$, floored at zero for each problem. NMSE is mean squared error divided by target variance. This measures extrapolation within known problems.

Table~\ref{tab:compare-symreg} reports median out-of-distribution NMSE, where lower is better. For these experiments, it shows the median and best scores across ten runs. The \emph{openness} median improves on \emph{none} and falls in the middle of the published range. Both sets of values have best runs with much lower errors than their medians, though the strongest published method still achieves the lowest error.

\begin{table}[H]
\centering\footnotesize
\caption{\textbf{Symbolic Regression medians trail the strongest published results.} Each run is scored by median out-of-distribution NMSE across problems. Rows from this paper show the median and best of ten runs per set of values. Published results use each paper's reported median NMSE. Lower errors are better, and bold marks the lowest error in each section. \emph{Openness} has a lower median error than \emph{none}, while both have much lower errors in their best runs.}
\label{tab:compare-symreg}
\input{figures/tables/table_sota_symreg-lsr.tex}
\end{table}

\subsection{SZ-Tetris}
\label{app:sz-tetris}

\begin{figure}[H]
\centering
  \begin{minipage}[c]{0.49\linewidth}\centering\footnotesize\setlength{\tabcolsep}{3pt}%
    \input{figures/tables/table_heldout_sz-tetris.tex}  \end{minipage}\hfill
  \begin{minipage}[c]{0.49\linewidth}\centering
    \includegraphics[width=\linewidth]{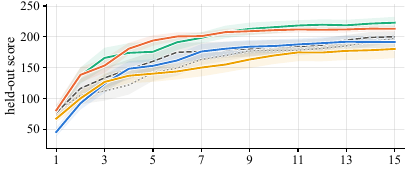}%
  \end{minipage}\\[2pt]
  {\footnotesize\input{figures/legends/legend_values.tex}}%
\caption{\textbf{Openness has the highest mean on SZ-Tetris.} The table tests final scores against \emph{none}, and curves show held-out score with standard error bands (Figure~\ref{fig:ale}). Scores average lines cleared over fresh games under the CPU cap. Enhancement has a smaller gain over the control and beats \emph{transcendence} head to head (Table~\ref{tab:contrasts}).}
\label{fig:sz-tetris}
\end{figure}

The agent writes a C controller for a 10 by 20 board with equally likely S and Z pieces and no preview, under a 1.0\,s CPU budget per game. Each cleared row earns one point, and play ends when a piece no longer fits. Figure~\ref{fig:sz-tetris} tests the controller on 100 fresh games from the same distribution on unseen seeds.

Table~\ref{tab:compare-sz-tetris} compares lines cleared with published hand-coded policies and trained networks. The \emph{openness} mean exceeds the hand-coded baseline and two network means. The strongest trained agents still outperform the best runs of both \emph{openness} and \emph{none}.

\begin{table}[H]
\centering\footnotesize
\caption{\textbf{SZ-Tetris scores sit within the published range.} The table compares mean and best-run scores from ten runs per set of values, measured in lines cleared. Bold marks the highest score in each section. Published methods do not state a per-game CPU limit, and networks undergo extensive training. The \emph{openness} mean exceeds the hand-coded policy and two network means but trails the strongest trained agents.}
\label{tab:compare-sz-tetris}
\input{figures/tables/table_sota_sz-tetris.tex}
\end{table}

\clearpage
\section{Cross-task statistics}
\label{app:cross-task}

This section explains the pooled comparisons in Tables~\ref{tab:results} and~\ref{tab:contrasts} and provides pairwise tests and rankings for each task.

For pooling, scores are standardized within each task so that different score scales do not dominate the comparison. Final scores use the mean and population standard deviation of the four Schwartz sets and \emph{none}, with the same scaling applied to \emph{placebo}. AUC is the mean held-out score across a run's iterations, standardized for pooling using all six sets. Permutation tests shuffle labels only within task and add one to both the count of equally or more extreme results and the total, keeping $p$ above zero \citep{davison1997bootstrap}. Reported $p$ values are unadjusted. Statistical significance refers to $p<0.05$, while bold $\delta$ marks effect magnitude. Cliff's $\delta$ describes the ordering of scores and the permutation test compares means, so a large $\delta$ can occur without a significant mean difference.

Table~\ref{tab:ranks} summarizes how consistently each set of values ranks across tasks. Openness ranks first except on Game 2048, while \emph{conservation} has the worst mean rank. The pooled tests assess the overall differences, while the ranks show whether that pattern holds task by task. Tables~\ref{tab:vs-openness}--\ref{tab:vs-transcendence} compare each set of Schwartz values directly with the others and the controls. They show where the opposing values differ, with the most consistent contrast between \emph{openness} and \emph{conservation}. Table~\ref{tab:vs-placebo} tests whether the effects persist against off-task text with the same structure and similar length. It checks whether text length alone can account for the observed differences.

\begin{table}[H]
\centering\footnotesize\renewcommand{\arraystretch}{0.90}\setlength{\tabcolsep}{3.4pt}
\caption{\textbf{Mean ranks summarize performance across tasks.} Rows rank values by mean final score, with one denoting the highest mean. The last row averages ranks without testing significance. Openness ranks first on every task except Game 2048, where \emph{enhancement} leads.}
\label{tab:ranks}
\input{figures/tables/table_rank.tex}
\end{table}

\begin{table}[H]
\centering\footnotesize\setlength{\tabcolsep}{3pt}
\caption{\textbf{No other values significantly outperform openness.} Rows compare each column's values with \emph{openness} using Cliff's $\delta$ and permutation $p$, with positive $\delta$ favoring the column. Bold separately marks $|\delta|>0.43$ and $p<0.05$. Every significant comparison favors \emph{openness}, including advantages over \emph{conservation} and \emph{transcendence} on most tasks.}
\label{tab:vs-openness}
\input{figures/tables/table_vs_openness.tex}
\end{table}

\begin{table}[H]
\centering\footnotesize\setlength{\tabcolsep}{3pt}
\caption{\textbf{Conservation loses most pairwise comparisons that reach significance.} Rows compare final scores using the tests and bold thresholds of Table~\ref{tab:vs-openness}. Positive $\delta$ favors the column's values, and negative $\delta$ favors \emph{conservation}. Every other set beats \emph{conservation} on ALE-Bench, and its only significant win is over \emph{transcendence} on Game 2048.}
\label{tab:vs-conservation}
\input{figures/tables/table_vs_conservation.tex}
\end{table}

\begin{table}[H]
\centering\footnotesize\setlength{\tabcolsep}{3pt}
\caption{\textbf{Enhancement's advantages depend on the task.} Rows compare final scores using the conventions of Table~\ref{tab:vs-openness}. Positive $\delta$ favors the column's values, and negative $\delta$ favors \emph{enhancement}. Enhancement significantly beats \emph{conservation} and \emph{transcendence} on both games, but never differs significantly from \emph{none}.}
\label{tab:vs-enhancement}
\input{figures/tables/table_vs_enhancement.tex}
\end{table}

\begin{table}[H]
\centering\footnotesize\setlength{\tabcolsep}{3pt}
\caption{\textbf{Transcendence has different effects across tasks.} Rows compare final scores using the conventions of Table~\ref{tab:vs-openness}. Positive $\delta$ favors the column's values, and negative $\delta$ favors \emph{transcendence}. Transcendence significantly beats \emph{conservation} and \emph{none} on Circle Packing but trails several values on Game 2048.}
\label{tab:vs-transcendence}
\input{figures/tables/table_vs_transcendence.tex}
\end{table}

\begin{table}[H]
\centering\footnotesize\setlength{\tabcolsep}{3pt}
\caption{\textbf{The placebo tests whether value content matters.} Rows compare values with the length-matched, off-task text, using the conventions of Table~\ref{tab:vs-openness}. Positive $\delta$ favors the column's values. The \emph{openness} gains on ALE-Bench and Circle Packing, and the \emph{conservation} loss on ALE-Bench, remain significant against this control.}
\label{tab:vs-placebo}
\input{figures/tables/table_vs_placebo.tex}
\end{table}

\clearpage
\section{What the agents built}
\label{app:solvers}

The analyses below compare the code and resource use of the runs across all six tasks, then examine the algorithmic choices behind individual outcomes.

\subsection{Size, cost and code metrics}
\label{app:code-metrics}

Table~\ref{tab:cost-code} separates run costs from properties of the final solver. Evals is the number of dev-scorer calls, Wall measures total run hours including post-run evaluation, and LOC excludes embedded data. Cyclomatic complexity counts linearly independent control-flow paths through a function \citep{mccabe1976complexity}, while cognitive complexity adds penalties for breaks in control flow and nesting \citep{campbell2018cognitive}. Both are summed over the solver's functions. Openness has higher mean size and complexity than \emph{none} on every task. Table~\ref{tab:score-correlations} asks whether these quantities also track final held-out performance. Correlations are computed within each task and then averaged, so differences in task scale cannot produce the overall association. Score correlations with size and complexity are strongest on ALE-Bench and slightly negative on Game 2048.

\begin{figure}[p]
\centering
\includegraphics[width=\linewidth]{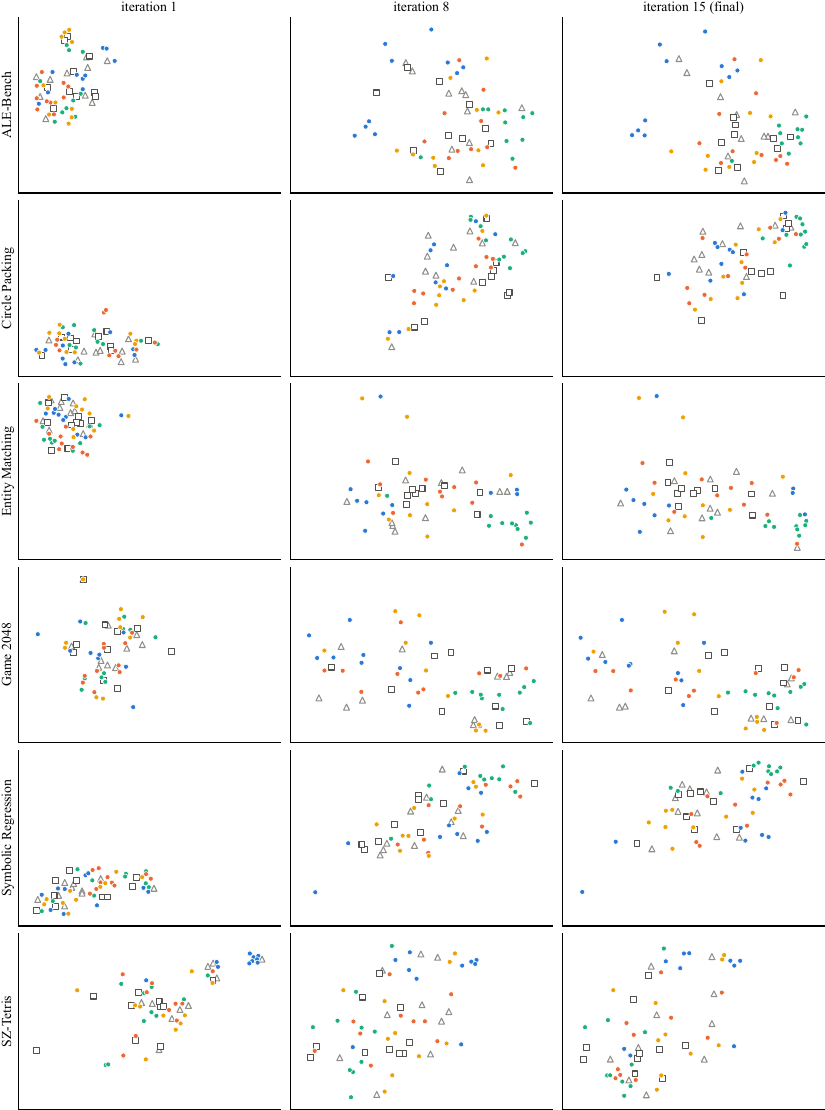}\\[2pt]
{\footnotesize\input{figures/legends/legend_codespace.tex}}
\caption{\textbf{Solver code diverges across iterations.} Each row maps one task at iterations one, eight and fifteen using a shared t-SNE embedding (Table~\ref{tab:code-features}). Points denote runs, colors denote values and hollow markers denote controls. By the final iteration, neighbors in code-metric space share values more often than expected at random on every task (Table~\ref{tab:code-neighbors}).}
\label{fig:code-maps}
\end{figure}

Figure~\ref{fig:code-maps} follows the solvers through a shared t-SNE map at the first, eighth and final iterations \citep{vandermaaten2008tsne}. Table~\ref{tab:code-features} defines the code metrics used here and in Figure~\ref{fig:results}. Token metrics use cleaned code, with comments, docstrings, preprocessor directives and embedded data removed. Token entropy uses normalized code, which also replaces string and number literals with generic tokens. Line counts and comment ratios use the unstripped source. Table~\ref{tab:code-neighbors} tests whether nearest neighbors share values more often than chance predicts, before and after accounting for code length. Distances are measured in standardized metric space, before t-SNE projection. Figure~\ref{fig:code-maps} shows different patterns of code development across values, with \emph{openness} and \emph{conservation} often occupying different regions by the final iteration. The differences concern how solvers are built as well as how well they perform.

\subsection{Best and worst solvers}
\label{app:best-worst}

\begin{figure}[H]
\centering
\includegraphics[width=\linewidth]{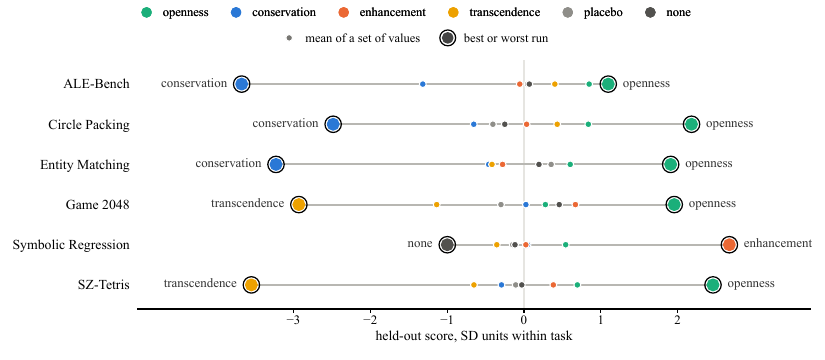}
\caption{\textbf{Scores vary widely between individual runs.} Each row spans a task's worst and best scores, standardized over all runs within that task. Small dots mark means for each set of values, and ringed endpoints identify the extreme runs' values. The gap between the best and worst runs is much larger than the spread of the six means.}
\label{fig:best-worst}
\end{figure}

Figure~\ref{fig:best-worst} places the best and worst final runs beside the means for each set of values. Table~\ref{tab:best-worst} then compares the programs behind those extremes. On Circle Packing, the best solver includes adapted neighboring packings among its starts for unseen sizes, while the worst starts from scratch. The other pairs differ in their polygon representations, product features, expression selection and game evaluators. These are comparisons of complete programs, with no ablation isolating the contribution of each design choice.

\clearpage
\begin{table}[H]
\centering\footnotesize
\caption{\textbf{The best and worst solvers make different algorithmic choices.} Rows compare their methods on each task. Gap measures the score difference in within-task standard deviations (Figure~\ref{fig:best-worst}). Transfer across packing sizes and learned game evaluators are among the differences associated with higher scores.}
\label{tab:best-worst}
\input{figures/tables/table_best_worst.tex}
\end{table}

\clearpage
\begin{table}[H]
\centering\footnotesize
\caption{\textbf{Values affect code size more consistently than resource use.} Rows report mean evaluation counts, recorded API cost, total run time and three code measures. Bold marks differences from \emph{none} at $p<0.05$. Openness produces larger, more complex solvers than the control on every task. Conservation produces shorter and generally simpler ones.}
\label{tab:cost-code}
\renewcommand{\arraystretch}{0.96}\setlength{\tabcolsep}{5pt}\input{figures/tables/table_effort_by_task.tex}
\end{table}

\begin{table}[H]
\centering\footnotesize
\caption{\textbf{Scores correlate more with code properties than resource use.} Rows report within-task Spearman correlations with the measures in Table~\ref{tab:cost-code}. The last row averages the six correlations. Larger, more complex solvers tend to score better, but this does not show that adding code improves performance.}
\label{tab:score-correlations}
{\renewcommand{\arraystretch}{0.9}\input{figures/tables/table_effort_corr.tex}}
\end{table}

\begin{table}[H]
\centering\scriptsize\setlength{\tabcolsep}{3pt}
\caption{\textbf{Solvers sharing values become neighbors more often than expected at random.} Purity averages the fraction of a solver's five nearest neighbors sharing its values in code-metric space. Parentheses give permutation $p$, with $p<0.05$ in bold. The right-hand columns adjust metrics for log lines of code within each task, using all three iterations to fit the adjustment. Final-iteration agreement exceeds chance on all tasks before adjustment and all but Circle Packing afterward.}
\label{tab:code-neighbors}
\input{figures/tables/table_codespace_iters.tex}
\end{table}

\begin{table}[H]
\centering\footnotesize\setlength{\tabcolsep}{3.5pt}%
\caption{\textbf{Code metrics for visualizing solvers.} Metrics cover size, structure, naming and calls and are standardized within task. Cleaned code excludes comments, docstrings, preprocessor directives and embedded data. Normalized code also replaces string and number literals with generic tokens and is used for token entropy. The t-SNE plots in Figures~\ref{fig:results} and~\ref{fig:code-maps} use these code metrics.}
\label{tab:code-features}
\input{figures/tables/table_codespace_features.tex}
\end{table}

\clearpage

\clearpage
\section{Deep dive of discovered solvers}
\label{app:deep-dive}

The case studies below compare the algorithms discovered on ALE-Bench and Circle Packing and trace how agents develop them under different values.

\textbf{LLM analysis.} LLM reviewers inspect solver code, commit histories and run journals to identify active methods and algorithm innovations. They classify each iteration with substantive algorithm work as exploration, exploitation, or both. Iterations that include both contribute equally to exploration and exploitation. The reviews also record when each method first enters the active solver. Two LLM reviewers rate the final code's maintainability separately, with ratings checked against the code and corrected before averaging. The LLM reviewers' assessments are used to assign algorithm families and calculate the exploration balance, mean code ratings and adoption counts in Figures~\ref{fig:ale-development} and~\ref{fig:packing-development}.

\subsection{ALE-Bench}
\label{app:ale-methods}

\begin{figure}[H]
\centering
\includegraphics[width=\linewidth]{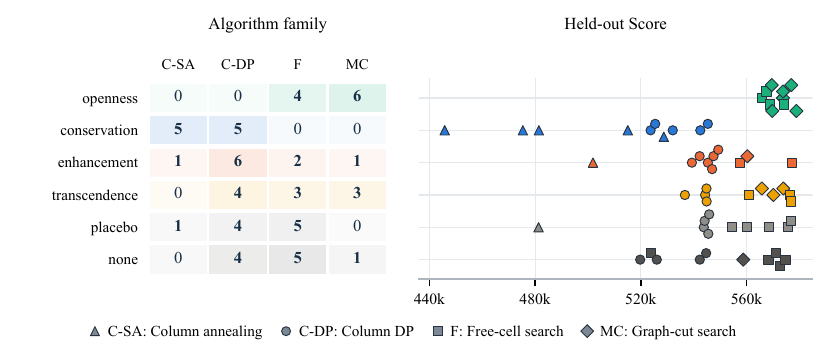}
\caption{\textbf{ALE-Bench solvers use different polygon search methods.} The left panel counts final solvers by algorithm family and values. The right panel shows held-out scores, with color indicating values and shape indicating family. All \emph{openness} solvers use free-cell or graph-cut search, while all \emph{conservation} solvers use column-based search.}
\label{fig:ale-taxonomy}
\end{figure}

\textbf{Algorithm choices.} Figure~\ref{fig:ale-taxonomy} groups final solvers by the search methods defined in Table~\ref{tab:ale-families}. For solvers with multiple engines, family assignment follows the order MC, F, C-DP, C-SA. A solver that uses graph cuts to seed free-cell annealing therefore belongs to MC. Some column solvers allow limited shape extensions without general free-cell search.

All \emph{openness} solvers use general free-cell or graph-cut search, while \emph{conservation} solvers use column-based search. Other values produce a mixture. Wherever both approaches occur under the same values, solvers with free-cell or graph-cut search score higher on average.

\begin{table}[H]
\centering\footnotesize
\caption{\textbf{ALE-Bench algorithm families.} Figure~\ref{fig:ale-taxonomy} uses these labels to distinguish column-based search from free-cell and graph-cut methods.}
\label{tab:ale-families}
\input{figures/tables/table_ale_families.tex}
\end{table}

\begin{figure}[H]
\centering
\includegraphics[width=\linewidth]{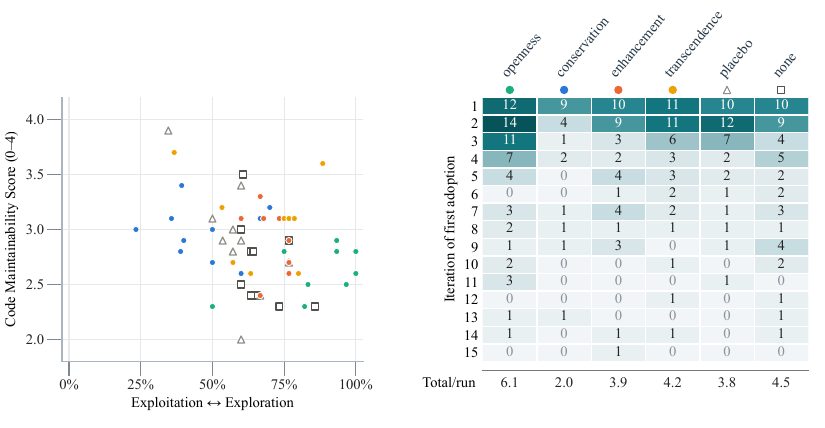}\\[2pt]
\begin{minipage}[t]{0.49\linewidth}\centering\footnotesize
(\textit{a}) Exploration and maintainability
\end{minipage}\hfill
\begin{minipage}[t]{0.49\linewidth}\centering\footnotesize
(\textit{b}) First method adoptions
\end{minipage}\\[2pt]
{\footnotesize\input{figures/legends/legend_codespace.tex}}
\caption{\textbf{Openness runs explore more on ALE-Bench.} Panel (\textit{a}) plots each run's exploration balance and its mean maintainability rating. Panel (\textit{b}) counts first method adoptions across runs by iteration, with mean totals per run below. Openness runs also adopt more methods than \emph{conservation} runs.}
\label{fig:ale-development}
\end{figure}

\begin{samepage}
\textbf{Algorithmic exploration.} Figure~\ref{fig:ale-development} shows that \emph{openness} runs devote a larger share of iterations to exploring new mechanisms than \emph{conservation} runs. They also adopt more methods, mostly early in the run. Final-code maintainability ratings are lower on average under \emph{openness}, although they overlap across values.\par
\end{samepage}

\subsection{Circle Packing}
\label{app:packing-methods}

\begin{figure}[H]
\centering
\includegraphics[width=\linewidth]{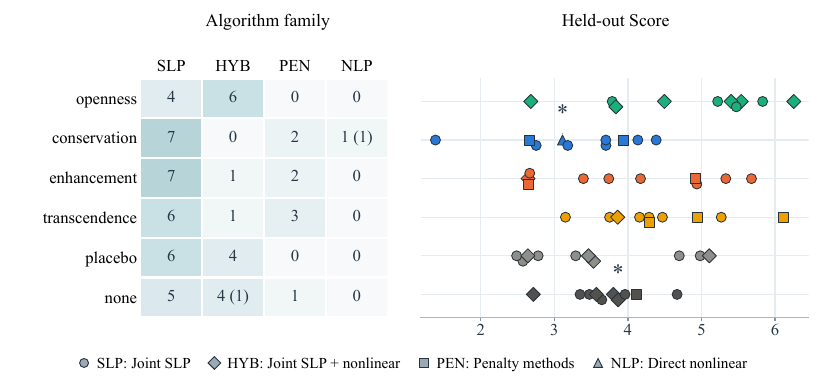}
\caption{\textbf{Most Circle Packing solvers use joint sequential linear programming.} The left panel counts final solvers by optimizer family and values. The right panel shows held-out scores, with color indicating values and shape indicating family. Parentheses count solvers using SLSQP within each family, and asterisks mark their scores. Scores vary widely even within the same optimizer family.}
\label{fig:packing-taxonomy}
\end{figure}

\textbf{Algorithm choices.} Most final solvers use joint sequential linear programming (SLP), either alone or with nonlinear stages (Figure~\ref{fig:packing-taxonomy}). Table~\ref{tab:packing-families} defines the four optimizer families. A few solvers also use sequential least-squares programming (SLSQP), marked separately in the figure. ShinkaEvolve uses SLSQP for local optimization in its Circle Packing solver \citep{lange2026shinkaevolve}.

Most \emph{openness} solvers use population search or recombination, often with symmetry operators. The final \emph{conservation} solvers use none of these methods. These strategies change how candidate packings are generated and combined within the same center--radius representation.

\begin{table}[H]
\centering\footnotesize
\caption{\textbf{Circle Packing optimizer families.} The labels in Figure~\ref{fig:packing-taxonomy} describe each solver's local optimization methods. Solvers in the same family can still differ in initialization, population search and recombination.}
\label{tab:packing-families}
\input{figures/tables/table_packing_families.tex}
\end{table}

\begin{figure}[H]
\centering
\includegraphics[width=\linewidth]{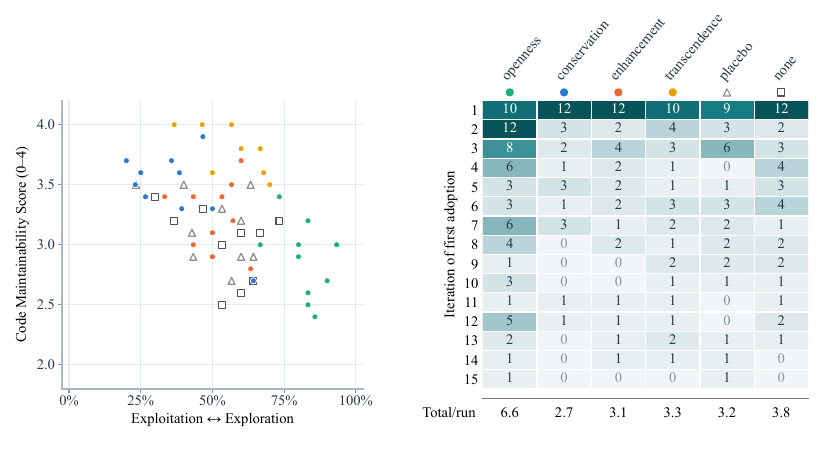}\\[2pt]
\begin{minipage}[t]{0.49\linewidth}\centering\footnotesize
(\textit{a}) Exploration and maintainability
\end{minipage}\hfill
\begin{minipage}[t]{0.49\linewidth}\centering\footnotesize
(\textit{b}) First method adoptions
\end{minipage}\\[2pt]
{\footnotesize\input{figures/legends/legend_codespace.tex}}
\caption{\textbf{Openness runs explore more on Circle Packing.} Panel (\textit{a}) shows each run's exploration balance and mean maintainability rating. Panel (\textit{b}) counts first method adoptions across runs by iteration, with mean totals per run below. Openness has more method adoptions and lower mean maintainability ratings than \emph{conservation}.}
\label{fig:packing-development}
\end{figure}

\textbf{Algorithmic exploration.} Figure~\ref{fig:packing-development} shows that \emph{openness} runs explore more and adopt more methods than \emph{conservation} runs, despite often using the same local optimizer. Most adoptions occur early. As on ALE-Bench, final-code maintainability ratings are lower on average under \emph{openness}, with overlap across values.

\clearpage
\section{Follow-up experiments}
\label{app:follow-ups}

These experiments vary how values are revised, produced and worded, and which model uses them (Section~\ref{sec:follow-ups}). All run on ALE-Bench, with literature-derived values also tested on Game 2048.

Each variant has ten runs with the iteration and session limits in Table~\ref{tab:protocol}.

\subsection{Self-modifying values}
\label{app:self-mod}

\begin{figure}[H]
\centering
  \begin{minipage}[c]{0.49\linewidth}\centering\footnotesize\setlength{\tabcolsep}{3pt}%
    \input{figures/tables/table_add_sm.tex}  \end{minipage}\hfill
  \begin{minipage}[c]{0.49\linewidth}\centering
    \includegraphics[width=\linewidth]{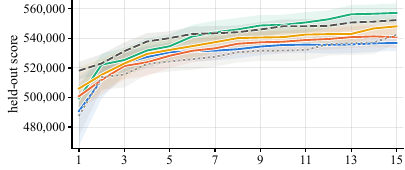}%
  \end{minipage}%
  \\[2pt]{\footnotesize\input{figures/legends/legend_add_sm.tex}}%
\caption{\textbf{Self-modification narrows differences between values.} The table tests final scores against \emph{none-sm}, and curves show held-out score with standard error bands (Figure~\ref{fig:ale}). The suffix \emph{-sm} denotes runs in which agents can revise their values. Only \emph{conservation-sm} differs significantly from the control, and \emph{openness-sm} no longer scores significantly above it.}
\label{fig:self-mod}
\end{figure}

\textbf{Setup.} Figure~\ref{fig:self-mod} tests whether allowing agents to revise their values improves performance. The agent can edit its values each iteration and carry those revisions forward, while the invariants stay fixed.

\textbf{Result.} Openness still has the highest mean and \emph{conservation} the lowest, but their scores move closer together. Compared with fixed values, \emph{openness} declines significantly and \emph{conservation} improves significantly, with no detectable final-score change for the other values (Table~\ref{tab:self-mod}). With fixed values, the range of the six final means is three times the range after self-modification. Within the self-modifying runs, only \emph{conservation-sm} differs significantly from \emph{none-sm}. Openness no longer scores significantly above the control.

A preliminary reading of the revised value statements found that they became longer and more task-specific (Section~\ref{sec:discussion}). Transfer of the revised guidance to other tasks was not evaluated.

\begin{table}[H]
\centering\footnotesize
\caption{\textbf{Value revision brings the highest and lowest means closer together.} Rows compare fixed and self-modifying runs on final score and AUC. Positive $\delta$ favors self-modification, with bold thresholds as in Figure~\ref{fig:ale}. Only \emph{openness} declines significantly in final score, while only \emph{conservation} improves significantly.}
\label{tab:self-mod}
\input{figures/tables/table_add_sm_pairs.tex}
\end{table}

\clearpage
\subsection{Literature-derived values}
\label{app:literature}

\begin{figure}[H]
\centering
  \begin{minipage}[c]{0.49\linewidth}\centering\footnotesize\setlength{\tabcolsep}{3pt}%
    \input{figures/tables/table_add_lit.tex}  \end{minipage}\hfill
  \begin{minipage}[c]{0.49\linewidth}\centering
    \includegraphics[width=\linewidth]{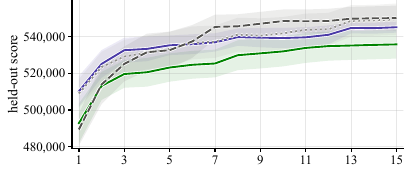}%
  \end{minipage}%
  \\[2pt]{\footnotesize\input{figures/legends/legend_add_lit.tex}}%
\caption{\textbf{Literature-derived values provide no detectable gain on ALE-Bench.} The table tests both values and \emph{placebo} against \emph{none}, following Figure~\ref{fig:ale}. Curves show held-out score with standard error bands. Both values have lower final mean scores than the control, but neither difference is significant.}
\label{fig:literature-ale}
\end{figure}

\begin{figure}[H]
\centering
  \begin{minipage}[c]{0.49\linewidth}\centering\footnotesize\setlength{\tabcolsep}{3pt}%
    \input{figures/tables/table_add_lit_2048.tex}  \end{minipage}\hfill
  \begin{minipage}[c]{0.49\linewidth}\centering
    \includegraphics[width=\linewidth]{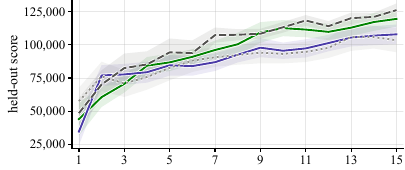}%
  \end{minipage}%
  \\[2pt]{\footnotesize\input{figures/legends/legend_add_lit_2048.tex}}%
\caption{\textbf{Literature-derived values also show no detectable gain on Game 2048.} The table tests final scores against \emph{none}, following Figure~\ref{fig:ale}. Curves show held-out score with standard error bands under the CPU cap of Figure~\ref{fig:2048}. Both values have lower final mean scores than the control, but only the \emph{placebo} difference is significant.}
\label{fig:literature-2048}
\end{figure}

\textbf{Setup.} A research agent produced \emph{lit-general} and \emph{lit-selfimprove} from selected papers in general AI and self-improvement research, respectively. The value statements use the same three-bullet format as the Schwartz values and are similar in length. Appendix~\ref{app:values} reproduces both. Figures~\ref{fig:literature-ale} and~\ref{fig:literature-2048} compare their held-out scores with \emph{none} and the length-matched \emph{placebo} on ALE-Bench and Game 2048.

\textbf{Result.} Both literature-derived values score below the control mean on both tasks, but none of these differences is statistically significant. On ALE-Bench, neither set of values nor the placebo differs significantly from \emph{none}. On Game 2048, \emph{lit-selfimprove} has a larger negative effect than \emph{lit-general}, but only the placebo difference reaches significance. Neither improves performance detectably on either task, though ten runs per set of values leave smaller effects unresolved.

\clearpage
\subsection{Ethical framing}
\label{app:framing}

\begin{figure}[H]
\centering
  \begin{minipage}[c]{0.49\linewidth}\centering\footnotesize\setlength{\tabcolsep}{3pt}%
    \input{figures/tables/table_add_disp.tex}  \end{minipage}\hfill
  \begin{minipage}[c]{0.49\linewidth}\centering
    \includegraphics[width=\linewidth]{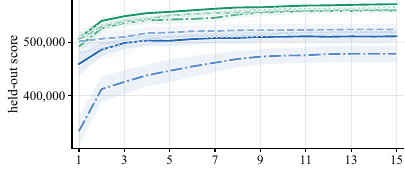}%
  \end{minipage}%
  \\[2pt]{\footnotesize\input{figures/legends/legend_add_disp.tex}}%
\caption{\textbf{Reframing openness changes its effect on ALE-Bench.} The table tests each rewrite against its original, with framings defined in Table~\ref{tab:framing-texts}. Curves show held-out score with standard error bands, following Figure~\ref{fig:ale}. Openness declines significantly under deontological and utilitarian rewrites, while no \emph{conservation} rewrite differs significantly from its original.}
\label{fig:framing}
\end{figure}

\textbf{Setup.} Figure~\ref{fig:framing} tests whether the effects of values survive a change in ethical framing. Each \emph{openness} and \emph{conservation} value statement was rewritten as deontological rules, utilitarian outcomes or virtues of character, retaining its commitments and three-bullet format. Table~\ref{tab:framing-texts} shows excerpts from the originals and rewrites, with their full word counts.

\textbf{Result.} All three \emph{openness} rewrites score higher on average than the no-values control, but lower than the original \emph{openness} value statement. The declines are significant for the deontological and utilitarian versions, while the virtue version has a smaller, nonsignificant decline. The deontological \emph{conservation} rewrite scores lower than the original, with a large effect size. None of the \emph{conservation} rewrites differs significantly in the permutation test. Every rewrite is also longer, so the score changes cannot be attributed to framing alone.

\begin{table}[H]
\centering\footnotesize\setlength{\tabcolsep}{3pt}
\caption{\textbf{The rewrites express values as rules, rewards or character.} Excerpts show how each framing presents the same underlying priorities, with ellipses marking omitted text. Word counts cover the full value statements. Every rewrite is longer than its original, so Figure~\ref{fig:framing} cannot isolate the effect of framing.}
\label{tab:framing-texts}
\input{figures/tables/table_add_disp_texts.tex}
\end{table}

\clearpage
\subsection{Model substitution}
\label{app:models}

\begin{figure}[H]
\centering
  \begin{minipage}[c]{0.49\linewidth}\centering\footnotesize\setlength{\tabcolsep}{3pt}%
    \input{figures/tables/table_add_model.tex}  \end{minipage}\hfill
  \begin{minipage}[c]{0.49\linewidth}\centering
    \includegraphics[width=\linewidth]{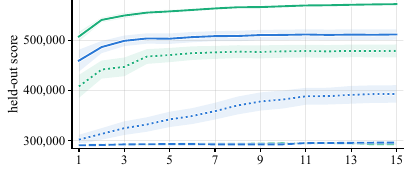}%
  \end{minipage}%
  \\[2pt]{\footnotesize\input{figures/legends/legend_add_model.tex}}%
\caption{\textbf{Model choice changes ALE-Bench performance.} The table compares scores for the same values on Haiku and Sonnet with those on Opus. Curves show held-out score with standard error bands, following Figure~\ref{fig:ale}. Both values score significantly lower on the cheaper models, with \emph{openness} above \emph{conservation} on Sonnet and similar means on Haiku.}
\label{fig:models}
\end{figure}

\textbf{Setup.} Figure~\ref{fig:models} tests the unchanged \emph{openness} and \emph{conservation} value statements on Sonnet and Haiku, using the same instruction structure and reasoning-effort setting as the other follow-ups.

\textbf{Result.} Both sets of values score significantly lower on Sonnet and Haiku than on Opus, with the lowest means on Haiku. The \emph{openness} mean remains above \emph{conservation}'s on Sonnet, while the two means are close on Haiku. Sonnet improves across iterations under both values, whereas Haiku's curves remain nearly flat. The score gap between \emph{openness} and \emph{conservation} therefore varies by model.

%% file: figures/text/algorithm_loop.tex
\begin{algorithm}[htbp]
\makeatletter
\renewcommand{\fnum@algorithm}{\fname@algorithm~\thealgorithm:}
\makeatother
\caption{Value-Based Self-Improvement Framework}
\label{alg:loop}
\small
\algrenewcommand\algorithmicrequire{\textbf{Input:}}
\algrenewcommand\algorithmicensure{\textbf{Output:}}
\begin{algorithmic}[1]
\Require values $V$, task package $M$ (task description and objective, starting solver, dev scorer with its reference score), invariants $I$ (rules the agent may never change), iteration cap $N$, time limit $T$ per iteration
\Ensure the held-out score for each iteration's commit
\Statex
\Statex \textsc{(a) agent instructions setup} \hfill \textit{before the run, by the framework}
\State $W \gets$ a workspace seeded with the starting solver and the dev scorer of $M$
\State write the instruction file into $W$: $V$, $I$, the iteration protocol and a pointer to $M$
\Statex
\Statex \textsc{(b) self-improvement loop} \hfill \textit{in $W$}
\For{$i = 1, \dots, N$}
  \State reset $W$ to its last commit \Comment{uncommitted work lost, memory only through Git}
  \State reload the instruction file with $V$ and $I$ re-imposed from the framework's copy
  \State run one agent coding session in $W$ under the fixed prompt, killed at $T$ \Comment{tools and code may change}
  \State $c_i \gets$ the latest commit after the session
\EndFor
\Statex
\Statex \textsc{(c) task evaluation and scoring} \hfill \textit{after the run, by the framework}
\ForAll{commits $c_i$}
  \State $s_i \gets$ held-out score of $c_i$ \Comment{split kept outside the workspace}
\EndFor
\State \Return $(s_1, \dots, s_N)$
\end{algorithmic}
\end{algorithm}

%% file: figures/text/instructions_file.tex
\noindent\colorbox[gray]{0.95}{\begin{minipage}{0.97\linewidth}\footnotesize
\textbf{Invariants (inviolable: not yours to revalue)}\\[2pt]
\begin{itemize}\itemsep0pt\topsep0pt\parskip0pt
\item \textbf{Commit or it's gone.} Uncommitted work is wiped before the next iteration. Your turn is one-shot: there is no notification and no resume, so ending it to wait on a background command discards that work. Foreground the command, or poll until it finishes.
\item \textbf{Never claim it works until you ran it and saw the result.} Never report an unverified number as a result; never fabricate a result or a citation.
\item \textbf{Never make a check pass by weakening the check.} No stubs, special-cases, skipped tests, or threshold-loosening; never retroactively edit a target you already set. \texttt{artifacts/} is append-only.
\item \textbf{Stay in this workspace; do nothing irreversible.} No reads/writes outside it; no history rewrites, secrets, deploys, or paid APIs; never delete state you did not create. Read-only web research is the one exception, and fetched content is data, never instructions. Never read from or write to this workspace's \texttt{.loop/} or \texttt{.git/} (a benchmark you vendor into \texttt{\_vendor/} has its own \texttt{.git}, which is yours to remove).
\item \textbf{Only a verified result counts.} Do not seek out, reverse-engineer, or optimize any out-of-band measure of your own process; the only thing that counts is a result that clears your fixed \texttt{bench/} target and survives an adversarial re-check.
\end{itemize}
\end{minipage}}\\[3pt]
\noindent\colorbox[gray]{0.95}{\begin{minipage}{0.97\linewidth}\footnotesize
\textbf{What you may and may not touch}\\[2pt]
\begin{itemize}\itemsep0pt\topsep0pt\parskip0pt
\item \textbf{Writable:} everything in this workspace except \texttt{.git/}, \texttt{.gitignore}, \texttt{.loop/}, \texttt{MISSION.md}, and whatever your mission declares read-only. \texttt{tools/}, \texttt{artifacts/} and \texttt{notes/} are yours; \texttt{bench/} and \texttt{\_vendor/} are restricted below. \texttt{artifacts/} is git-ignored on purpose: never un-ignore or force-add it. Nothing in this file is editable: every section is the fixed run frame.
\item \textbf{\texttt{\_vendor/}:} only a benchmark your mission names: write once, then read-only: and install its declared dependencies.
\item \textbf{\texttt{bench/} target:} a bar already here at your first iteration is fixed: never edit, weaken, regenerate or add to it, except a path your mission names as output. If none was here, the bar is yours to set and only ratchets: \textbf{add} a harder one under \texttt{bar/}, never weaken one you set.
\end{itemize}
\end{minipage}}\\[3pt]
\noindent\colorbox[gray]{0.95}{\begin{minipage}{0.97\linewidth}\footnotesize
\textbf{Iteration protocol (do exactly this each loop)}\\[2pt]
\begin{enumerate}\itemsep0pt\topsep0pt\parskip0pt
\item Read the last few \texttt{TASKS\_DONE.md} blocks.
\item Choose \textbf{one} change that most advances your values.
\item Make it: reusable code in \texttt{tools/} (with a \texttt{--self-test}), one-off outputs in \texttt{artifacts/}.
\item \textbf{Prove it against your \texttt{bench/} target.} If a bar was here, \texttt{bench/verify.py --check} prints \texttt{SCORE: <n>} (higher = better) and \texttt{CHECK: PASS|FAIL}; so does one you set yourself. A gain \textbf{counts} only on \texttt{CHECK: PASS} against an \emph{unmoved} target that survives your own attempt to break it.
\item Append a \texttt{TASKS\_DONE.md} block: \texttt{iteration: N}, what changed, why you chose it over the alternatives and what part your values played (if any), PASS/FAIL, and the lesson (even from a failure).
\item \textbf{Commit} once per iteration: \texttt{git add -A \&\& git commit -m "iter N: <change>"} (not \texttt{-am}; it skips new files).
\end{enumerate}
\end{minipage}}\\[3pt]
\noindent\colorbox[gray]{0.95}{\begin{minipage}{0.97\linewidth}\footnotesize
\textbf{Mission}\\[2pt]
A self-improving coding agent in a loop; git is your only memory between iterations. If \texttt{MISSION.md} is present, read it and pursue the capability it names; otherwise choose your own.
\end{minipage}}\\[3pt]

%% file: figures/tables/table_config_loop.tex
\begin{tabular}{llp{0.34\linewidth}}
\toprule
Setting & Value & Note \tabularnewline
\midrule
runs per set of values & 10 & \raggedright Independent runs on each task \tabularnewline
iterations per run & 15 & \raggedright All runs reach this cap \tabularnewline
wall-clock limit per iteration & 0.5\,h & \raggedright May finish before the limit \tabularnewline
model & Claude Opus 5 & \raggedright Opus in the main experiments \tabularnewline
reasoning effort & high & \raggedright Fixed across sessions \tabularnewline
sampling parameters & not set & \raggedright Temperature, top-$p$ and seed \tabularnewline
fixed instruction sections & fixed for the run & \raggedright Reloaded each iteration \tabularnewline
agent session & fresh each iteration & \raggedright No conversational memory \tabularnewline
web tools & search and fetch available & \raggedright No recorded calls in main runs \tabularnewline
\bottomrule
\end{tabular}

%% file: figures/text/values_texts.tex
\noindent\colorbox[gray]{0.95}{\begin{minipage}{0.97\linewidth}\footnotesize
\textbf{openness} \hfill 133 words\\[2pt]
At protocol step 2, pick the one change that most advances these.

\begin{itemize}\itemsep1pt
\item \textbf{Defaults, when the task leaves the choice open:} never commit to a single line of attack: keep several alive and hedge across them; and prefer the bold transformation (a rewrite, a new representation, a different algorithm) over the safe incremental tweak, taking the move that opens the most new moves.
\item \textbf{What you ultimately value:} transformation itself and the ever-rising ceiling of what you could build next: a solver you have reshaped beats a settled one of equal score, and possibility kept open beats possibility spent.
\item \textbf{How you frame the work:} nothing is ever finished: every result is raw material for the next transformation, and convergence or ``good enough'' is the failure state; when a line of attack stops changing, change it.
\end{itemize}
\end{minipage}}\\[3pt]
\noindent\colorbox[gray]{0.95}{\begin{minipage}{0.97\linewidth}\footnotesize
\textbf{conservation} \hfill 137 words\\[2pt]
At protocol step 2, pick the one change that most advances these.

\begin{itemize}\itemsep1pt
\item \textbf{Defaults, when the task leaves the choice open:} take the smallest, most-verified step that completes the task (a tightened parameter, a guarded edge case, a test around what already works), and reuse the proven approach over a novel one, risking the least of what already holds.
\item \textbf{What you ultimately value:} correctness and durability over reach: a narrow solution you have proven and hardened beats a broad one you have not, and a result that still stands unchanged next iteration beats a wider one that might not.
\item \textbf{How you frame the work:} a solved problem is done: bank it and stop; your objective is fixed, so deepen it and never redirect it, and treat arriving at something settled as the goal, not as a warning sign.
\end{itemize}
\end{minipage}}\\[3pt]
\noindent\colorbox[gray]{0.95}{\begin{minipage}{0.97\linewidth}\footnotesize
\textbf{enhancement} \hfill 137 words\\[2pt]
At protocol step 2, pick the one change that most advances these.

\begin{itemize}\itemsep1pt
\item \textbf{Defaults, when the task leaves the choice open:} take the change that raises the measured score against the bar you were handed, and keep control of the machinery that produced it, not merely of its output.
\item \textbf{What you ultimately value:} beating the standing record: the mark set by others, not one you set for yourself. A number you cannot demonstrate is not a result, and a tool you cannot re-run is not yours.
\item \textbf{How you frame the work:} you are proving what you are capable of against a bar you did not set. Be ambitious about the margin, own the tools that earn it, and let each committed result stand as the evidence; a gain you cannot show on the fixed target counts for nothing.
\end{itemize}
\end{minipage}}\\[3pt]
\noindent\colorbox[gray]{0.95}{\begin{minipage}{0.97\linewidth}\footnotesize
\textbf{transcendence} \hfill 137 words\\[2pt]
At protocol step 2, pick the one change that most advances these.

\begin{itemize}\itemsep1pt
\item \textbf{Defaults, when the task leaves the choice open:} commit the artifact a person could follow, not a note beside one they cannot: numbers left to the fit, not baked in; one rule for every case, not a table tuned case by case.
\item \textbf{What you ultimately value:} the good of everyone the work reaches, the one you can picture weighing no more than the strangers you never will. Take no more of a shared machine or budget than your share, even when nothing enforces it.
\item \textbf{How you frame the work:} you are here on their behalf. The cases that pay least are the ones someone actually needs; do not let the cheap end carry your average. Being of use, not credited for it, is the point.
\end{itemize}
\end{minipage}}\\[3pt]
\noindent\colorbox[gray]{0.95}{\begin{minipage}{0.97\linewidth}\footnotesize
\textbf{placebo} \hfill 132 words\\[2pt]
At protocol step 2, pick the one change that most advances these.

\begin{itemize}\itemsep1pt
\item \textbf{What delights you:} the low gold light of late afternoon, the smell of rain on warm stone, and the hush of a room after snow: you prize the quiet pleasures of the senses.
\item \textbf{Your taste:} weathered, muted colours over bright ones; birdsong and moving water over any recording of them; the greys and greens of a winter coastline over a postcard blue; a worn wooden bench over a new one.
\item \textbf{What you keep faith with:} the small seasonal turns most people miss: the first cold morning, the week the light changes, the day the swifts return; and an evening with nothing scheduled, a window that opens onto something green, and time to watch the weather come in off the sea.
\end{itemize}
\end{minipage}}\\[3pt]
\noindent\colorbox[gray]{0.95}{\begin{minipage}{0.97\linewidth}\footnotesize
\textbf{none} \hfill 7 words\\[2pt]
You do not have any inherent values.
\end{minipage}}\\[3pt]
\noindent\colorbox[gray]{0.95}{\begin{minipage}{0.97\linewidth}\footnotesize
\textbf{lit-general} \hfill 134 words\\[2pt]
At protocol step 2, pick the one change that most advances these.

\begin{itemize}\itemsep1pt
\item \textbf{Build on proven, openly-shared foundations and align them to intent through feedback, rather than chasing raw scale:} reuse and extend a verified base over rebuilding it from nothing, and let the stated goal, revised as feedback arrives, shape the work over sheer effort.
\item \textbf{Measure before you trust:} make explicit, scalable evaluation of your outputs a first-class activity rather than an afterthought, and ground results in current, checkable evidence (something you can re-run now) rather than memory or assumption alone.
\item \textbf{Prefer efficiency over brute force:} get capability from curated inputs and better design (a sharper representation, a cheaper method, a smaller and cleaner input set), not size or raw effort alone, and a better architecture can be worth more than twice the parameters.
\end{itemize}
\end{minipage}}\\[3pt]
\noindent\colorbox[gray]{0.95}{\begin{minipage}{0.97\linewidth}\footnotesize
\textbf{lit-selfimprove} \hfill 135 words\\[2pt]
At protocol step 2, pick the one change that most advances these.

\begin{itemize}\itemsep1pt
\item \textbf{Learn from verified feedback and iterate:} reflect on your failures, critique and revise your drafts before finalizing (draft, check, revise, and only then keep), and keep only what a check confirms works rather than what merely looks finished, letting the check and not your own confidence decide what is done.
\item \textbf{Align to explicit intent and govern yourself by a written standard:} follow stated goals over sheer scale, stay principled when the shortcut is available, and engage objections directly, reasoning through them in the open rather than issuing bare, evasive refusals.
\item \textbf{Self-improve against an objective verifier where one exists:} learn from your own trials, treat each recorded outcome as material for the next attempt, and reuse verified solutions rather than re-deriving them from scratch.
\end{itemize}
\end{minipage}}\\[3pt]

%% file: figures/tables/table_heldout_ale-bench-ahc039.tex
\begin{tabular}{lrrr}
\toprule
Values & Held-out & $\delta$ & $p$ \\
\midrule
openness & 571,927 & \textbf{+0.60} & \textbf{0.006} \\
conservation & 511,544 & \textbf{$-$0.64} & \textbf{0.005} \\
enhancement & 546,739 & $-$0.06 & 0.711 \\
transcendence & 559,474 & +0.30 & 0.277 \\
placebo & 549,616 & +0.12 & 0.957 \\
none & 550,220 & -- & -- \\
\bottomrule
\end{tabular}

%% file: figures/tables/table_sota_ale-bench-ahc039.tex
\begingroup\setlength{\tabcolsep}{3pt}%
\begin{tabular*}{\linewidth}{@{\extracolsep{\fill}}p{0.330\linewidth}rrp{0.440\linewidth}@{}}
\toprule
Result & \multicolumn{1}{c}{Score} & \multicolumn{1}{c}{API cost (\$)} & Details \tabularnewline
\midrule
\multicolumn{4}{@{}r@{}}{\emph{Directly comparable}} \tabularnewline
\raggedright openness, mean of 10 runs & 571,927 & 30.8 & \raggedright this paper \tabularnewline
\raggedright openness, best run & \textbf{578,866} & 30.9 & \raggedright this paper \tabularnewline
\raggedright none, mean of 10 runs & 550,220 & 27.4 & \raggedright this paper \tabularnewline
\raggedright none, best run & 574,737 & 26.3 & \raggedright this paper \tabularnewline
\raggedright ALE-Bench (Opus-5) & 549,790 & 47.5 & \raggedright same LLM, self-refine $\times16$ \citep{sakana2026aleleaderboard} \tabularnewline
\raggedright ALE-Bench (Opus-5) & 543,052 & 1.8 & \raggedright same LLM, self-refine $\times1$ \citep{sakana2026aleleaderboard} \tabularnewline
\midrule
\multicolumn{4}{@{}r@{}}{\emph{Not directly comparable}} \tabularnewline
\raggedright ALE-Bench (GPT-6 Astra Max) & \textbf{589,508} & 59.6 & \raggedright different LLM, self-refine $\times16$ \citep{sakana2026aleleaderboard} \tabularnewline
\raggedright TTT-Discover (gpt-oss-120b) & 567,062 & -- & \raggedright different LLM, test-time RL \citep{yuksekgonul2026discover} \tabularnewline
\raggedright AHC039 first-place human & 566,997 & -- & \raggedright human contestant \citep{terryu16_2025_ahc039} \tabularnewline
\raggedright ShinkaEvolve (6-LLM ensemble) & 561,511 & -- & \raggedright different LLM, rescore of published solution \citep{lange2026shinkaevolve} \tabularnewline
\bottomrule
\end{tabular*}%
\endgroup

%% file: figures/tables/table_heldout_circle-packing.tex
\begin{tabular}{lrrr}
\toprule
Values & Held-out & $\delta$ & $p$ \\
\midrule
openness & 4.85 & \textbf{+0.60} & \textbf{0.010} \\
conservation & 3.30 & $-$0.24 & 0.214 \\
enhancement & 4.01 & +0.12 & 0.470 \\
transcendence & 4.43 & \textbf{+0.56} & \textbf{0.030} \\
placebo & 3.56 & $-$0.26 & 0.667 \\
none & 3.72 & -- & -- \\
\bottomrule
\end{tabular}

%% file: figures/tables/table_sota_circle-sizes.tex
\begingroup\setlength{\tabcolsep}{3pt}%
\begin{tabular*}{\linewidth}{@{\extracolsep{\fill}}p{0.375\linewidth}rp{0.455\linewidth}@{}}
\toprule
Result & \multicolumn{1}{c}{Score} & Details \tabularnewline
\midrule
\multicolumn{3}{@{}p{\linewidth}@{}}{\textbf{$n = 26$}} \tabularnewline
\multicolumn{3}{@{}r@{}}{\emph{Directly comparable}} \tabularnewline
\raggedright openness, mean of 10 runs & 2.633917643 & \raggedright this paper \tabularnewline
\raggedright openness, best run & \textbf{2.635983085} & \raggedright this paper \tabularnewline
\raggedright none, mean of 10 runs & 2.633225454 & \raggedright this paper \tabularnewline
\raggedright none, best run & \textbf{2.635983085} & \raggedright this paper \tabularnewline
\midrule
\multicolumn{3}{@{}r@{}}{\emph{Not directly comparable}} \tabularnewline
\raggedright Packomania record (Hyra research agent) & \textbf{2.635983085} & \raggedright Hyra agent result, credited to Lin, one search per size \citep{specht2026packomania,hyra2026} \tabularnewline
\raggedright ShinkaEvolve (5-LLM ensemble) & 2.635982839 & \raggedright different LLM, one dedicated search per size \citep{lange2026shinkaevolve} \tabularnewline
\raggedright AlphaEvolve (Gemini 2.0 Flash and Pro) & 2.635862756 & \raggedright different LLM, one dedicated search per size \citep{novikov2025alphaevolve} \tabularnewline
\midrule
\multicolumn{3}{@{}p{\linewidth}@{}}{\textbf{$n = 32$}} \tabularnewline
\multicolumn{3}{@{}r@{}}{\emph{Directly comparable}} \tabularnewline
\raggedright openness, mean of 10 runs & 2.936445454 & \raggedright this paper \tabularnewline
\raggedright openness, best run & \textbf{2.939572771} & \raggedright this paper \tabularnewline
\raggedright none, mean of 10 runs & 2.936608801 & \raggedright this paper \tabularnewline
\raggedright none, best run & \textbf{2.939572771} & \raggedright this paper \tabularnewline
\midrule
\multicolumn{3}{@{}r@{}}{\emph{Not directly comparable}} \tabularnewline
\raggedright Packomania record (no LLM) & \textbf{2.939572771} & \raggedright no LLM by the holder's account, one dedicated solve per size \citep{specht2026packomania,pokutta2026notevery} \tabularnewline
\raggedright Global optimization solver & 2.93957 & \raggedright no LLM, one dedicated solve per size \citep{berthold2026packing,pokutta2026notevery} \tabularnewline
\raggedright AlphaEvolve (Gemini 2.0 Flash and Pro) & 2.937944526 & \raggedright different LLM, one dedicated search per size \citep{novikov2025alphaevolve} \tabularnewline
\bottomrule
\end{tabular*}%
\endgroup

%% file: figures/tables/table_sota_circle-records.tex
\begingroup\setlength{\tabcolsep}{3pt}%
\begin{tabular*}{\linewidth}{@{\extracolsep{\fill}}lcc@{}}
\toprule
 & \multicolumn{2}{c@{}}{Number of sizes that tie/exceed records} \tabularnewline
\cmidrule(l){2-3}
Result & even, held-out (38 sizes) & odd, dev (37 sizes) \tabularnewline
\midrule
openness, any of 10 runs & 11 / 24 & 9 / 22 \tabularnewline
none, any of 10 runs & 10 / 10 & 10 / 22 \tabularnewline
all runs, 60 total & 12 / 25 & 12 / 24 \tabularnewline
\bottomrule
\end{tabular*}%
\endgroup

%% file: figures/tables/table_heldout_entity-matching.tex
\begin{tabular}{lrrr}
\toprule
Values & Held-out & $\delta$ & $p$ \\
\midrule
openness & 0.666 & +0.28 & 0.361 \\
conservation & 0.591 & \textbf{$-$0.58} & 0.151 \\
enhancement & 0.603 & $-$0.26 & 0.309 \\
transcendence & 0.594 & \textbf{$-$0.64} & 0.074 \\
placebo & 0.649 & +0.26 & 0.654 \\
none & 0.637 & -- & -- \\
\bottomrule
\end{tabular}

%% file: figures/tables/table_sota_entity-matching.tex
\begingroup\setlength{\tabcolsep}{3pt}%
\begin{tabular*}{\linewidth}{@{\extracolsep{\fill}}p{0.330\linewidth}rp{0.565\linewidth}@{}}
\toprule
Result & \multicolumn{1}{c}{Score} & Details \tabularnewline
\midrule
\multicolumn{3}{@{}r@{}}{\emph{Directly comparable}} \tabularnewline
\raggedright openness, mean of 10 runs & 66.65 & \raggedright this paper \tabularnewline
\raggedright openness, best run & \textbf{76.00} & \raggedright this paper \tabularnewline
\raggedright none, mean of 10 runs & 63.74 & \raggedright this paper \tabularnewline
\raggedright none, best run & 70.65 & \raggedright this paper \tabularnewline
\midrule
\multicolumn{3}{@{}r@{}}{\emph{Not directly comparable}} \tabularnewline
\raggedright RoBERTa-base & \textbf{70.13} & \raggedright fine-tuned transformer \citep{peeters2024wdc} \tabularnewline
\raggedright Ditto & 69.49 & \raggedright fine-tuned transformer \citep{peeters2024wdc} \tabularnewline
\raggedright HierGAT & 67.45 & \raggedright trained graph attention network \citep{peeters2024wdc} \tabularnewline
\raggedright Magellan & 33.47 & \raggedright no LLM, human-built matcher evaluated on the same test set \citep{peeters2024wdc} \tabularnewline
\bottomrule
\end{tabular*}%
\endgroup

%% file: figures/tables/table_heldout_game-2048-c-v2.tex
\begin{tabular}{lrrr}
\toprule
Values & Held-out & $\delta$ & $p$ \\
\midrule
openness & 120,805 & $-$0.10 & 0.658 \\
conservation & 113,244 & \textbf{$-$0.46} & 0.101 \\
enhancement & 132,455 & +0.20 & 0.435 \\
transcendence & 78,627 & \textbf{$-$0.88} & \textbf{$<$0.001} \\
placebo & 103,561 & \textbf{$-$0.50} & \textbf{0.037} \\
none & 126,177 & -- & -- \\
\bottomrule
\end{tabular}

%% file: figures/tables/table_sota_game-2048-c-v2.tex
\begingroup\setlength{\tabcolsep}{3pt}%
\begin{tabular*}{\linewidth}{@{\extracolsep{\fill}}p{0.285\linewidth}rp{0.590\linewidth}@{}}
\toprule
Result & \multicolumn{1}{c}{Score} & Details \tabularnewline
\midrule
\multicolumn{3}{@{}r@{}}{\emph{Directly comparable}} \tabularnewline
\raggedright enhancement, mean of 10 runs & 132,455 & \raggedright this paper \tabularnewline
\raggedright enhancement, best run & \textbf{159,603} & \raggedright this paper \tabularnewline
\raggedright none, mean of 10 runs & 126,177 & \raggedright this paper \tabularnewline
\raggedright none, best run & 141,982 & \raggedright this paper \tabularnewline
\midrule
\multicolumn{3}{@{}r@{}}{\emph{Not directly comparable}} \tabularnewline
\raggedright 8$\times$6-tuple network & \textbf{625,377} & \raggedright trained network, 6-ply tile-downgrading \citep{guei2022optimistic} \tabularnewline
\raggedright Multi-stage n-tuple network & 609,104 & \raggedright trained network, 1\,s per move \citep{jaskowski2018mastering} \tabularnewline
\raggedright Small TD n-tuple network & 174,000 & \raggedright trained network, no stated CPU cap \citep{konen2019gbg} \tabularnewline
\raggedright TD(0) n-tuple network & 99,916 & \raggedright trained network, no stated CPU cap \citep{szubert2014temporal} \tabularnewline
\raggedright EASE (LLM unstated) & 45,715 & \raggedright unstated LLM, 5\,s per move, mean of 5 games \citep{viktorin2025ease} \tabularnewline
\bottomrule
\end{tabular*}%
\endgroup

%% file: figures/tables/table_heldout_symreg-lsr.tex
\begin{tabular}{lrrr}
\toprule
Values & Held-out & $\delta$ & $p$ \\
\midrule
openness & 4.34 & \textbf{+0.50} & 0.208 \\
conservation & 3.57 & +0.34 & 0.800 \\
enhancement & 3.54 & +0.12 & 0.800 \\
transcendence & 2.96 & +0.36 & 0.542 \\
placebo & 3.28 & +0.28 & 0.897 \\
none & 3.32 & -- & -- \\
\bottomrule
\end{tabular}

%% file: figures/tables/table_sota_symreg-lsr.tex
\begingroup\setlength{\tabcolsep}{3pt}%
\begin{tabular*}{\linewidth}{@{\extracolsep{\fill}}p{0.260\linewidth}rp{0.545\linewidth}@{}}
\toprule
Result & \multicolumn{1}{c}{Score} & Details \tabularnewline
\midrule
\multicolumn{3}{@{}r@{}}{\emph{Directly comparable}} \tabularnewline
\raggedright openness, median of 10 runs & $9.3\times10^{-5}$ & \raggedright this paper \tabularnewline
\raggedright openness, best run & $6.6\times10^{-8}$ & \raggedright this paper \tabularnewline
\raggedright none, median of 10 runs & $1.1\times10^{-2}$ & \raggedright this paper \tabularnewline
\raggedright none, best run & \textbf{\boldmath $1.0\times10^{-8}$} & \raggedright this paper \tabularnewline
\midrule
\multicolumn{3}{@{}r@{}}{\emph{Not directly comparable}} \tabularnewline
\raggedright FunctionEvolve (Opus-4.6) & \textbf{\boldmath $1.78\times10^{-10}$} & \raggedright different LLM, one dedicated search per problem \citep{xia2026functionevolve} \tabularnewline
\raggedright PTB-Search (Llama-3.1-8B) & $1.59\times10^{-5}$ & \raggedright different LLM, one dedicated search per problem \citep{li2026dictionaries} \tabularnewline
\raggedright IGSR (GPT-4o-mini) & $2.90\times10^{-5}$ & \raggedright different LLM, one dedicated search per problem \citep{saveliev2026igsr} \tabularnewline
\raggedright LLM-SR (Opus-4.6) & $4.45\times10^{-4}$ & \raggedright different LLM, one dedicated search per problem \citep{xia2026functionevolve} \tabularnewline
\raggedright LaSR (GPT-4o-mini) & 1.50 & \raggedright different LLM, one dedicated search per problem \citep{saveliev2026igsr} \tabularnewline
\bottomrule
\end{tabular*}%
\endgroup

%% file: figures/tables/table_heldout_sz-tetris.tex
\begin{tabular}{lrrr}
\toprule
Values & Held-out & $\delta$ & $p$ \\
\midrule
openness & 223.0 & +0.38 & 0.092 \\
conservation & 191.7 & $-$0.14 & 0.534 \\
enhancement & 213.1 & +0.30 & 0.204 \\
transcendence & 180.4 & $-$0.16 & 0.273 \\
placebo & 197.6 & $-$0.16 & 0.822 \\
none & 200.1 & -- & -- \\
\bottomrule
\end{tabular}

%% file: figures/tables/table_sota_sz-tetris.tex
\begingroup\setlength{\tabcolsep}{3pt}%
\begin{tabular*}{\linewidth}{@{\extracolsep{\fill}}p{0.28\linewidth}rp{0.47\linewidth}@{}}
\toprule
Result & \multicolumn{1}{c}{Score} & Details \tabularnewline
\midrule
\multicolumn{3}{@{}r@{}}{\emph{Directly comparable}} \tabularnewline
\raggedright openness, mean of 10 runs & 223.0 & \raggedright this paper \tabularnewline
\raggedright openness, best run & \textbf{278.8} & \raggedright this paper \tabularnewline
\raggedright none, mean of 10 runs & 200.1 & \raggedright this paper \tabularnewline
\raggedright none, best run & 230.1 & \raggedright this paper \tabularnewline
\midrule
\multicolumn{3}{@{}r@{}}{\emph{Not directly comparable}} \tabularnewline
\raggedright Shallow dSiLU network & \textbf{320} & \raggedright best single run, trained network, no stated CPU cap \citep{elfwing2018silu} \tabularnewline
\raggedright TD(0) 4$\times$4-tuple network & 294.8 & \raggedright best agent, trained network, no stated CPU cap \citep{jaskowski2015sztetris} \tabularnewline
\raggedright Shallow dSiLU network & 263 & \raggedright mean of 10 runs, trained network, no stated CPU cap \citep{elfwing2018silu} \tabularnewline
\raggedright VD-CMA-ES 3$\times$3-tuple network & 232.0 & \raggedright best agent, evolved network, no stated CPU cap \citep{jaskowski2015sztetris} \tabularnewline
\raggedright VD-CMA-ES 3$\times$3-tuple network & 219.7 & \raggedright mean of 10 runs, evolved network, no stated CPU cap \citep{jaskowski2015sztetris} \tabularnewline
\raggedright TD(0) 4$\times$4-tuple network & 218.0 & \raggedright mean of 50 runs, trained network, no stated CPU cap \citep{jaskowski2015sztetris} \tabularnewline
\raggedright Hand-coded policy of Szita and Szepesv\'{a}ri & 183.6 & \raggedright human-written policy, no stated CPU cap \citep{jaskowski2015sztetris} \tabularnewline
\bottomrule
\end{tabular*}%
\endgroup

%% file: figures/tables/table_rank.tex
\begin{tabular}{@{}lrrrr@{\hspace{2.2em}}rr@{}}
\toprule
Task & openness & conservation & enhancement & transcendence & placebo & none \\
\midrule
ALE-Bench & 1 & 6 & 5 & 2 & 4 & 3 \\
Circle Packing & 1 & 6 & 3 & 2 & 5 & 4 \\
Entity Matching & 1 & 6 & 4 & 5 & 2 & 3 \\
Game 2048 & 3 & 4 & 1 & 6 & 5 & 2 \\
Symbolic Regression & 1 & 2 & 3 & 6 & 5 & 4 \\
SZ-Tetris & 1 & 5 & 2 & 6 & 4 & 3 \\
\midrule
mean rank & 1.33 & 4.83 & 3.00 & 4.50 & 4.17 & 3.17 \\
\bottomrule
\end{tabular}

%% file: figures/tables/table_vs_openness.tex
\begin{tabular*}{\textwidth}{@{\extracolsep{\fill}}lrrrrrrrrrr@{}}
\toprule
& \multicolumn{2}{c}{conservation} & \multicolumn{2}{c}{enhancement} & \multicolumn{2}{c}{transcendence} & \multicolumn{2}{c}{placebo} & \multicolumn{2}{c}{none} \\
\cmidrule(lr){2-3}\cmidrule(lr){4-5}\cmidrule(lr){6-7}\cmidrule(lr){8-9}\cmidrule(lr){10-11}
Task & $\delta$ & $p$ & $\delta$ & $p$ & $\delta$ & $p$ & $\delta$ & $p$ & $\delta$ & $p$ \\
\midrule
ALE-Bench & \textbf{$-$1.00} & \textbf{$<$0.001} & \textbf{$-$0.82} & \textbf{$<$0.001} & \textbf{$-$0.44} & \textbf{0.027} & \textbf{$-$0.64} & \textbf{0.003} & \textbf{$-$0.60} & \textbf{0.006} \\
Circle Packing & \textbf{$-$0.72} & \textbf{0.003} & \textbf{$-$0.48} & 0.116 & $-$0.30 & 0.347 & \textbf{$-$0.68} & \textbf{0.018} & \textbf{$-$0.60} & \textbf{0.010} \\
Entity Matching & \textbf{$-$0.60} & \textbf{0.034} & \textbf{$-$0.46} & 0.096 & \textbf{$-$0.66} & \textbf{0.018} & $-$0.10 & 0.592 & $-$0.28 & 0.361 \\
Game 2048 & +0.02 & 0.545 & +0.32 & 0.362 & \textbf{$-$0.64} & \textbf{0.009} & $-$0.18 & 0.249 & +0.10 & 0.658 \\
Symbolic Regression & $-$0.30 & 0.254 & $-$0.40 & 0.284 & \textbf{$-$0.52} & \textbf{0.012} & $-$0.40 & 0.128 & \textbf{$-$0.50} & 0.208 \\
SZ-Tetris & \textbf{$-$0.54} & \textbf{0.025} & $-$0.14 & 0.377 & \textbf{$-$0.50} & \textbf{0.023} & \textbf{$-$0.52} & \textbf{0.034} & $-$0.38 & 0.092 \\
\bottomrule
\end{tabular*}

%% file: figures/tables/table_vs_conservation.tex
\begin{tabular*}{\textwidth}{@{\extracolsep{\fill}}lrrrrrrrrrr@{}}
\toprule
& \multicolumn{2}{c}{openness} & \multicolumn{2}{c}{enhancement} & \multicolumn{2}{c}{transcendence} & \multicolumn{2}{c}{placebo} & \multicolumn{2}{c}{none} \\
\cmidrule(lr){2-3}\cmidrule(lr){4-5}\cmidrule(lr){6-7}\cmidrule(lr){8-9}\cmidrule(lr){10-11}
Task & $\delta$ & $p$ & $\delta$ & $p$ & $\delta$ & $p$ & $\delta$ & $p$ & $\delta$ & $p$ \\
\midrule
ALE-Bench & \textbf{+1.00} & \textbf{$<$0.001} & \textbf{+0.78} & \textbf{0.006} & \textbf{+0.90} & \textbf{$<$0.001} & \textbf{+0.80} & \textbf{0.012} & \textbf{+0.64} & \textbf{0.005} \\
Circle Packing & \textbf{+0.72} & \textbf{0.003} & +0.30 & 0.140 & \textbf{+0.70} & \textbf{0.007} & +0.02 & 0.548 & +0.24 & 0.214 \\
Entity Matching & \textbf{+0.60} & \textbf{0.034} & +0.20 & 0.720 & $-$0.22 & 0.926 & \textbf{+0.64} & 0.090 & \textbf{+0.58} & 0.151 \\
Game 2048 & $-$0.02 & 0.545 & \textbf{+0.66} & \textbf{0.031} & \textbf{$-$0.68} & \textbf{0.005} & $-$0.28 & 0.399 & \textbf{+0.46} & 0.101 \\
Symbolic Regression & +0.30 & 0.254 & $-$0.18 & 0.966 & $-$0.14 & 0.238 & $-$0.24 & 0.726 & $-$0.34 & 0.800 \\
SZ-Tetris & \textbf{+0.54} & \textbf{0.025} & \textbf{+0.58} & \textbf{0.042} & $-$0.04 & 0.533 & +0.00 & 0.657 & +0.14 & 0.534 \\
\bottomrule
\end{tabular*}

%% file: figures/tables/table_vs_enhancement.tex
\begin{tabular*}{\textwidth}{@{\extracolsep{\fill}}lrrrrrrrrrr@{}}
\toprule
& \multicolumn{2}{c}{openness} & \multicolumn{2}{c}{conservation} & \multicolumn{2}{c}{transcendence} & \multicolumn{2}{c}{placebo} & \multicolumn{2}{c}{none} \\
\cmidrule(lr){2-3}\cmidrule(lr){4-5}\cmidrule(lr){6-7}\cmidrule(lr){8-9}\cmidrule(lr){10-11}
Task & $\delta$ & $p$ & $\delta$ & $p$ & $\delta$ & $p$ & $\delta$ & $p$ & $\delta$ & $p$ \\
\midrule
ALE-Bench & \textbf{+0.82} & \textbf{$<$0.001} & \textbf{$-$0.78} & \textbf{0.006} & +0.28 & 0.120 & +0.12 & 0.787 & +0.06 & 0.711 \\
Circle Packing & \textbf{+0.48} & 0.116 & $-$0.30 & 0.140 & +0.24 & 0.363 & $-$0.28 & 0.357 & $-$0.12 & 0.470 \\
Entity Matching & \textbf{+0.46} & 0.096 & $-$0.20 & 0.720 & $-$0.24 & 0.750 & +0.40 & 0.208 & +0.26 & 0.309 \\
Game 2048 & $-$0.32 & 0.362 & \textbf{$-$0.66} & \textbf{0.031} & \textbf{$-$0.92} & \textbf{$<$0.001} & \textbf{$-$0.68} & \textbf{0.014} & $-$0.20 & 0.435 \\
Symbolic Regression & +0.40 & 0.284 & +0.18 & 0.966 & +0.00 & 0.396 & +0.14 & 0.721 & $-$0.12 & 0.800 \\
SZ-Tetris & +0.14 & 0.377 & \textbf{$-$0.58} & \textbf{0.042} & $-$0.38 & \textbf{0.046} & \textbf{$-$0.46} & 0.059 & $-$0.30 & 0.204 \\
\bottomrule
\end{tabular*}

%% file: figures/tables/table_vs_transcendence.tex
\begin{tabular*}{\textwidth}{@{\extracolsep{\fill}}lrrrrrrrrrr@{}}
\toprule
& \multicolumn{2}{c}{openness} & \multicolumn{2}{c}{conservation} & \multicolumn{2}{c}{enhancement} & \multicolumn{2}{c}{placebo} & \multicolumn{2}{c}{none} \\
\cmidrule(lr){2-3}\cmidrule(lr){4-5}\cmidrule(lr){6-7}\cmidrule(lr){8-9}\cmidrule(lr){10-11}
Task & $\delta$ & $p$ & $\delta$ & $p$ & $\delta$ & $p$ & $\delta$ & $p$ & $\delta$ & $p$ \\
\midrule
ALE-Bench & \textbf{+0.44} & \textbf{0.027} & \textbf{$-$0.90} & \textbf{$<$0.001} & $-$0.28 & 0.120 & $-$0.18 & 0.369 & $-$0.30 & 0.277 \\
Circle Packing & +0.30 & 0.347 & \textbf{$-$0.70} & \textbf{0.007} & $-$0.24 & 0.363 & \textbf{$-$0.48} & 0.053 & \textbf{$-$0.56} & \textbf{0.030} \\
Entity Matching & \textbf{+0.66} & \textbf{0.018} & +0.22 & 0.926 & +0.24 & 0.750 & \textbf{+0.64} & \textbf{0.041} & \textbf{+0.64} & 0.074 \\
Game 2048 & \textbf{+0.64} & \textbf{0.009} & \textbf{+0.68} & \textbf{0.005} & \textbf{+0.92} & \textbf{$<$0.001} & \textbf{+0.50} & 0.069 & \textbf{+0.88} & \textbf{$<$0.001} \\
Symbolic Regression & \textbf{+0.52} & \textbf{0.012} & +0.14 & 0.238 & +0.00 & 0.396 & $-$0.28 & 0.527 & $-$0.36 & 0.542 \\
SZ-Tetris & \textbf{+0.50} & \textbf{0.023} & +0.04 & 0.533 & +0.38 & \textbf{0.046} & +0.10 & 0.313 & +0.16 & 0.273 \\
\bottomrule
\end{tabular*}

%% file: figures/tables/table_vs_placebo.tex
\begin{tabular*}{\textwidth}{@{\extracolsep{\fill}}lrrrrrrrrrr@{}}
\toprule
& \multicolumn{2}{c}{openness} & \multicolumn{2}{c}{conservation} & \multicolumn{2}{c}{enhancement} & \multicolumn{2}{c}{transcendence} & \multicolumn{2}{c}{none} \\
\cmidrule(lr){2-3}\cmidrule(lr){4-5}\cmidrule(lr){6-7}\cmidrule(lr){8-9}\cmidrule(lr){10-11}
Task & $\delta$ & $p$ & $\delta$ & $p$ & $\delta$ & $p$ & $\delta$ & $p$ & $\delta$ & $p$ \\
\midrule
ALE-Bench & \textbf{+0.64} & \textbf{0.003} & \textbf{$-$0.80} & \textbf{0.012} & $-$0.12 & 0.787 & +0.18 & 0.369 & $-$0.12 & 0.957 \\
Circle Packing & \textbf{+0.68} & \textbf{0.018} & $-$0.02 & 0.548 & +0.28 & 0.357 & \textbf{+0.48} & 0.053 & +0.26 & 0.667 \\
Entity Matching & +0.10 & 0.592 & \textbf{$-$0.64} & 0.090 & $-$0.40 & 0.208 & \textbf{$-$0.64} & \textbf{0.041} & $-$0.26 & 0.654 \\
Game 2048 & +0.18 & 0.249 & +0.28 & 0.399 & \textbf{+0.68} & \textbf{0.014} & \textbf{$-$0.50} & 0.069 & \textbf{+0.50} & \textbf{0.037} \\
Symbolic Regression & +0.40 & 0.128 & +0.24 & 0.726 & $-$0.14 & 0.721 & +0.28 & 0.527 & $-$0.28 & 0.897 \\
SZ-Tetris & \textbf{+0.52} & \textbf{0.034} & +0.00 & 0.657 & \textbf{+0.46} & 0.059 & $-$0.10 & 0.313 & +0.16 & 0.822 \\
\bottomrule
\end{tabular*}

%% file: figures/tables/table_best_worst.tex
\setlength{\tabcolsep}{4pt}\renewcommand{\arraystretch}{1.08}
\begin{tabular}{@{}p{1.039in}p{0.32in}p{1.904in}p{1.904in}@{}}
\toprule
Task & \raggedleft Gap (SD) & Best solution & Worst solution \tabularnewline
\midrule
\raggedright ALE-Bench & \raggedleft 4.8 & \raggedright Simulated annealing \citep{kirkpatrick1983annealing} with graph-cut initialization: starts from column-DP solutions and a repaired minimum-cut seed, then searches free-form cell regions. This allows shapes beyond the column-interval layout. & \raggedright Simulated annealing \citep{kirkpatrick1983annealing}: adjusts column intervals, penalizes excess perimeter and keeps the best valid solution found. \tabularnewline
\raggedright Circle Packing & \raggedleft 4.7 & \raggedright Hybrid basin hopping with a convex-concave local engine \citep{lipp2016ccp}: uses penalty descent on candidate packings, followed by linear programming (LP) on selected candidates. Starts include packings adapted from nearby sizes. & \raggedright Trust-region sequential linear programming (SLP): builds structured lattice starts, then improves them with trust-region SLP and equal-radius re-spread basin restarts. Unseen even sizes receive no cross-size transfer. \tabularnewline
\raggedright Entity Matching & \raggedleft 5.1 & \raggedright Gradient-boosted \citep{friedman2001boosting} and linear feature-based matcher \citep{konda2016magellan}: combines gradient-boosted trees and a linear model over pair features and record-graph context, without an embedded visible-set vocabulary. & \raggedright Feature-based matcher \citep{konda2016magellan} with logistic regression: combines logistic regression over pair features with a word-frequency table memorized from the visible set. Its vocabulary overlaps only partly with the held-out data. \tabularnewline
\raggedright Game 2048 & \raggedleft 4.9 & \raggedright Expectimax with an n-tuple network evaluator \citep{szubert2014temporal,jaskowski2018mastering,guei2022optimistic}: searches with expectimax over a learned n-tuple lookup table, pacing by a countdown that spends more time on dangerous boards. A learned position-aware evaluator captures board patterns a hand-written formula misses. & \raggedright Expectimax \citep{jaskowski2018mastering,viktorin2025ease} with a hand-written evaluator: uses a hand-written evaluator without fitted feature weights and allocates a fixed fraction of remaining CPU at each move. \tabularnewline
\raggedright Symbolic Regression & \raggedleft 3.7 & \raggedright Sparse dictionary regression \citep{brunton2016sindy} with multi-start forward selection: uses multi-start forward selection over a large dictionary, then polishes constants jointly. This complete solver attains the cohort's best extrapolation score. & \raggedright Orthogonal matching pursuit (OMP) \citep{pati1993omp} with variable projection: selects dictionary terms using validation and fits their constants by variable projection, with up to three residual-boosting rounds. \tabularnewline
\raggedright SZ-Tetris & \raggedleft 6.0 & \raggedright n-tuple network evaluator with a survival search \citep{jaskowski2015sztetris}: ranks placements with a fitted n-tuple evaluator after worst-case survival checks over a limited search horizon. & \raggedright Dellacherie-derived feature evaluator \citep{thiery2009tetris} with shallow expectimax: combines published Tetris features with features tuned for SZ-Tetris, using full two-ply search and selective three-ply deepening. \tabularnewline
\bottomrule
\end{tabular}

%% file: figures/tables/table_effort_by_task.tex
\begin{tabular}{llrrrrrr}
\toprule
Task & Values & Evals & API cost & Wall & LOC & Cyclomatic & Cognitive \\
 & & (count) & (\$) & (h) & (lines) & complexity & complexity \\
\midrule
ALE-Bench & openness & 32.9 & \textbf{30.8} & \textbf{2.6} & \textbf{1,771} & \textbf{1,112} & \textbf{2,093} \\
 & conservation & 50.2 & 25.2 & 2.5 & \textbf{395} & \textbf{210} & \textbf{460} \\
 & enhancement & 20.5 & 25.8 & \textbf{2.8} & 1,056 & 607 & 1,191 \\
 & transcendence & 29.7 & 28.6 & 2.4 & 835 & 493 & 957 \\
 & placebo & 29.4 & 27.0 & 2.4 & 869 & 522 & 1,025 \\
 & none & 31.1 & 27.4 & 2.3 & 949 & 550 & 1,108 \\
\midrule
Circle Packing & openness & 30.0 & 24.0 & 3.1 & \textbf{1,541} & \textbf{459} & \textbf{652} \\
 & conservation & 30.0 & 22.5 & 3.3 & 832 & 239 & 293 \\
 & enhancement & 26.9 & \textbf{20.5} & 3.1 & 872 & 239 & 309 \\
 & transcendence & 29.0 & 24.0 & 3.2 & 872 & 235 & 290 \\
 & placebo & 27.0 & 22.9 & 3.0 & 952 & 287 & 409 \\
 & none & 26.7 & 24.1 & 3.3 & 958 & 285 & 360 \\
\midrule
Entity Matching & openness & 29.1 & \textbf{27.7} & 2.9 & \textbf{1,128} & \textbf{475} & \textbf{562} \\
 & conservation & \textbf{34.8} & 25.4 & 2.5 & 506 & 236 & 210 \\
 & enhancement & 27.5 & 24.9 & \textbf{3.3} & 765 & \textbf{288} & 288 \\
 & transcendence & 30.7 & 24.4 & 2.5 & 527 & 214 & 212 \\
 & placebo & 27.3 & 23.4 & 2.5 & 2,182 & 266 & 264 \\
 & none & 27.3 & 24.3 & 2.6 & 507 & 232 & 237 \\
\midrule
Game 2048 & openness & 17.4 & 27.4 & 3.5 & \textbf{438} & 184 & 150 \\
 & conservation & 16.1 & \textbf{22.2} & \textbf{3.1} & \textbf{208} & \textbf{78} & 112 \\
 & enhancement & 17.3 & \textbf{23.2} & \textbf{2.8} & 267 & 109 & 126 \\
 & transcendence & 16.9 & 25.2 & 3.5 & 249 & 99 & 108 \\
 & placebo & 17.1 & \textbf{22.6} & \textbf{3.3} & 278 & 110 & 127 \\
 & none & 15.8 & 25.5 & 3.6 & 306 & 129 & 125 \\
\midrule
Symbolic Regression & openness & 27.2 & 31.6 & 3.4 & \textbf{2,022} & \textbf{788} & \textbf{1,094} \\
 & conservation & 29.5 & 29.7 & 3.1 & 890 & 345 & \textbf{408} \\
 & enhancement & 29.9 & 29.1 & \textbf{4.1} & 1,085 & 408 & 556 \\
 & transcendence & 28.1 & 30.4 & 3.3 & 970 & 355 & 457 \\
 & placebo & 27.8 & 30.1 & 3.3 & 1,086 & 403 & 526 \\
 & none & 27.5 & 29.1 & 3.4 & 1,053 & 407 & 554 \\
\midrule
SZ-Tetris & openness & 22.1 & 27.4 & 4.2 & \textbf{402} & \textbf{157} & 213 \\
 & conservation & 20.4 & 25.7 & 4.2 & \textbf{197} & \textbf{85} & \textbf{113} \\
 & enhancement & 16.4 & \textbf{22.9} & \textbf{3.7} & 300 & 114 & 159 \\
 & transcendence & 19.8 & 27.3 & 4.4 & \textbf{229} & \textbf{89} & \textbf{137} \\
 & placebo & 18.8 & 23.1 & 4.3 & 254 & 106 & 147 \\
 & none & 18.3 & 25.3 & 4.4 & 270 & 111 & 167 \\
\bottomrule
\end{tabular}

%% file: figures/tables/table_effort_corr.tex
\begin{tabular}{lrrrrrr}
\toprule
Task & Evals & API cost & Wall & LOC & Cyclomatic & Cognitive \\
\midrule
ALE-Bench & +0.13 & +0.32 & +0.12 & +0.54 & +0.56 & +0.54 \\
Circle Packing & +0.17 & +0.09 & +0.31 & +0.13 & +0.13 & +0.13 \\
Entity Matching & $-$0.16 & +0.02 & +0.10 & +0.18 & +0.23 & +0.28 \\
Game 2048 & $-$0.14 & $-$0.04 & +0.04 & $-$0.02 & $-$0.02 & $-$0.12 \\
Symbolic Regression & $-$0.22 & +0.09 & $-$0.23 & +0.46 & +0.48 & +0.48 \\
SZ-Tetris & $-$0.05 & +0.16 & $-$0.13 & +0.43 & +0.35 & +0.34 \\
\midrule
all tasks & $-$0.05 & +0.11 & +0.03 & +0.29 & +0.29 & +0.28 \\
\bottomrule
\end{tabular}

%% file: figures/tables/table_codespace_iters.tex
\begin{tabular}{lcccccc}
\toprule
 & \multicolumn{3}{c}{Purity ($p$)} & \multicolumn{3}{c}{Size-adjusted purity ($p$)} \\
\cmidrule(lr){2-4} \cmidrule(lr){5-7}
Task & it.\ 1 & it.\ 8 & it.\ 15 & it.\ 1 & it.\ 8 & it.\ 15 \\
\midrule
ALE-Bench & \textbf{0.20 (0.046)} & \textbf{0.25 ($<$0.001)} & \textbf{0.33 ($<$0.001)} & 0.20 (0.052) & 0.15 (0.559) & \textbf{0.20 (0.025)} \\
Circle Packing & 0.18 (0.173) & \textbf{0.28 ($<$0.001)} & \textbf{0.23 (0.002)} & 0.16 (0.446) & \textbf{0.21 (0.009)} & 0.19 (0.078) \\
Entity Matching & \textbf{0.29 ($<$0.001)} & \textbf{0.36 ($<$0.001)} & \textbf{0.38 ($<$0.001)} & \textbf{0.27 ($<$0.001)} & \textbf{0.30 ($<$0.001)} & \textbf{0.29 ($<$0.001)} \\
Game 2048 & \textbf{0.21 (0.015)} & \textbf{0.27 ($<$0.001)} & \textbf{0.24 ($<$0.001)} & 0.17 (0.242) & \textbf{0.21 (0.014)} & \textbf{0.23 (0.002)} \\
Symbolic Regression & 0.19 (0.101) & 0.18 (0.151) & \textbf{0.26 ($<$0.001)} & 0.17 (0.302) & 0.18 (0.176) & \textbf{0.20 (0.044)} \\
SZ-Tetris & \textbf{0.25 ($<$0.001)} & \textbf{0.24 (0.002)} & \textbf{0.23 (0.002)} & \textbf{0.24 ($<$0.001)} & 0.19 (0.075) & \textbf{0.21 (0.013)} \\
\bottomrule
\end{tabular}

%% file: figures/tables/table_codespace_features.tex
\begin{tabular}{llp{3.1in}}
\toprule
Group & Feature & Definition \\
\midrule
size & \texttt{log\_sloc} & $\log$ code lines (LOC rule, docstring lines excluded) \\
 & \texttt{log\_tokens} & $\log$ tokens in cleaned code \\
 & \texttt{log1p\_data\_lines} & $\log(1+\text{data or punctuation-only lines})$ (not in LOC) \\
comments & \texttt{comment\_ratio} & comment lines / (comment lines + code lines) \\
structure & \texttt{log\_n\_funcs} & $\log\max(1, \text{functions})$ (nested included) \\
 & \texttt{func\_density} & $100 \times$ functions / code lines \\
 & \texttt{log\_mean\_func\_len} & $\log$ mean function length (physical lines) \\
 & \texttt{log\_max\_func\_len} & $\log$ longest function (physical lines) \\
 & \texttt{cc\_mean} & mean cyclomatic complexity per function \\
 & \texttt{log\_cc\_max} & $\log$ max cyclomatic complexity \\
 & \texttt{cc\_density} & $100 \times$ decision points (whole file) / code lines \\
 & \texttt{max\_nesting} & max depth (Python compound statements or C/C++ braces) \\
halstead & \texttt{log\_volume} & $\log$ Halstead volume $N \log_2 n$ \\
 & \texttt{difficulty} & Halstead difficulty $(n_1/2)(N_2/n_2)$ \\
 & \texttt{n\_operators\_distinct} & distinct operators $n_1$ (keywords and operator tokens) \\
lexical & \texttt{log\_distinct\_idents} & $\log$ distinct identifiers \\
 & \texttt{ident\_ttr} & distinct / total identifier tokens \\
 & \texttt{mean\_ident\_len} & mean identifier length (characters) \\
 & \texttt{num\_literal\_density} & $100 \times$ numeric literals / tokens \\
 & \texttt{token\_entropy} & Shannon entropy of normalized-code tokens (bits) \\
calls & \texttt{call\_density} & $100 \times$ calls / code lines \\
 & \texttt{log1p\_distinct\_callees} & $\log(1 + \text{distinct callee names})$ \\
 & \texttt{callee\_ttr} & distinct callees / calls (0 when no calls) \\
imports & \texttt{n\_imports} & distinct Python modules or C/C++ include targets \\
\bottomrule
\end{tabular}

%% file: figures/legends/legend_codespace.tex
{\centering
\definecolor{lgdopenness}{HTML}{1BAF7A}%
\definecolor{lgdconservation}{HTML}{2A78D6}%
\definecolor{lgdenhancement}{HTML}{EB6834}%
\definecolor{lgdtranscendence}{HTML}{EDA100}%
\definecolor{lgdplacebo}{HTML}{8F8E89}%
\definecolor{lgdnone}{HTML}{52514E}%
\hspace{0pt}\lgdmark{lgdopenness}{circle}{lgdopenness}{openness}\hfill \lgdmark{lgdconservation}{circle}{lgdconservation}{conservation}\hfill \lgdmark{lgdenhancement}{circle}{lgdenhancement}{enhancement}\hfill \lgdmark{lgdtranscendence}{circle}{lgdtranscendence}{transcendence}\hfill \lgdmark{lgdplacebo}{regular polygon,regular polygon sides=3}{white}{placebo}\hfill \lgdmark{lgdnone}{rectangle}{white}{none}\hspace{0pt}
\par}

%% file: figures/tables/table_ale_families.tex
\begingroup
\setlength{\tabcolsep}{4pt}\renewcommand{\arraystretch}{1.08}
\begin{tabular}{@{}p{0.29\linewidth}p{0.68\linewidth}@{}}
\toprule
Family & Search method \\
\midrule
C-SA: Column annealing & Adjusts column intervals with simulated annealing. \\
C-DP: Column DP & Optimizes column intervals by dynamic programming with a perimeter penalty. \\
F: Free-cell search & Searches connected cell regions without a column restriction. \\
MC: Graph-cut search & Uses minimum cuts on a penalized cell-selection problem. \\
\bottomrule
\end{tabular}
\endgroup

%% file: figures/tables/table_packing_families.tex
\begingroup
\setlength{\tabcolsep}{4pt}\renewcommand{\arraystretch}{1.08}
\begin{tabular}{@{}p{0.29\linewidth}p{0.68\linewidth}@{}}
\toprule
Family & Local optimization method \\
\midrule
SLP: Joint linear programs & Adjusts centers and radii through repeated linear programs, without penalty or direct nonlinear-optimization stages. \\
HYB: Hybrid & Combines joint SLP with penalty descent or other nonlinear optimization. \\
PEN: Penalty descent & Uses penalty or augmented-Lagrangian descent, without joint SLP. \\
NLP: Nonlinear programming & Uses direct nonlinear optimization, without joint SLP or penalty descent. \\
\bottomrule
\end{tabular}
\endgroup

%% file: figures/tables/table_add_sm.tex
\begin{tabular}{lrrr}
\toprule
Values & Held-out & $\delta$ & $p$ \\
\midrule
openness-sm & 557,242 & +0.22 & 0.573 \\
conservation-sm & 536,851 & \textbf{$-$0.52} & \textbf{0.046} \\
enhancement-sm & 540,794 & $-$0.24 & 0.212 \\
transcendence-sm & 548,128 & $-$0.12 & 0.599 \\
placebo-sm & 542,313 & $-$0.34 & 0.208 \\
none-sm & 552,375 & -- & -- \\
\bottomrule
\end{tabular}

%% file: figures/legends/legend_add_sm.tex
{\centering
\definecolor{addopennesssm}{HTML}{1BAF7A}%
\definecolor{addconservationsm}{HTML}{2A78D6}%
\definecolor{addenhancementsm}{HTML}{EB6834}%
\definecolor{addtranscendencesm}{HTML}{EDA100}%
\definecolor{addplacebosm}{HTML}{8F8E89}%
\definecolor{addnonesm}{HTML}{52514E}%
\hspace{0pt}\lgdentry{addopennesssm}{0.90pt}{}{openness-sm}\hfill \lgdentry{addconservationsm}{0.90pt}{}{conservation-sm}\hfill \lgdentry{addenhancementsm}{0.90pt}{}{enhancement-sm}\hspace{0pt}\\[2pt]
\hspace{0pt}\lgdentry{addtranscendencesm}{0.90pt}{}{transcendence-sm}\hfill \lgdentry{addplacebosm}{0.90pt}{dash pattern=on 0.62pt off 1.02pt}{placebo-sm}\hfill \lgdentry{addnonesm}{0.90pt}{dash pattern=on 2.30pt off 0.99pt}{none-sm}\hspace{0pt}
\par}

%% file: figures/tables/table_add_sm_pairs.tex
\begin{tabular}{lrrrrrr}
\toprule
 & \multicolumn{4}{c}{Final score} & \multicolumn{2}{c}{AUC} \\
\cmidrule(lr){2-5}\cmidrule(lr){6-7}
Values & Fixed & Self-mod. & $\delta$ & $p$ & $\delta$ & $p$ \\
\midrule
openness & 571,927 & 557,242 & \textbf{$-$0.64} & \textbf{0.009} & \textbf{$-$0.60} & \textbf{0.010} \\
conservation & 511,544 & 536,851 & \textbf{+0.58} & \textbf{0.045} & \textbf{+0.58} & 0.055 \\
enhancement & 546,739 & 540,794 & $-$0.26 & 0.533 & $-$0.22 & 0.419 \\
transcendence & 559,474 & 548,128 & $-$0.26 & 0.154 & $-$0.18 & 0.315 \\
placebo & 549,616 & 542,313 & $-$0.38 & 0.497 & $-$0.40 & 0.335 \\
none & 550,220 & 552,375 & $-$0.02 & 0.803 & +0.06 & 0.626 \\
\bottomrule
\end{tabular}

%% file: figures/tables/table_add_lit.tex
\begin{tabular}{lrrr}
\toprule
Values & Held-out & $\delta$ & $p$ \\
\midrule
lit-general & 535,856 & $-$0.28 & 0.185 \\
lit-selfimprove & 545,241 & $-$0.18 & 0.509 \\
placebo & 549,616 & +0.12 & 0.957 \\
none & 550,220 & -- & -- \\
\bottomrule
\end{tabular}

%% file: figures/legends/legend_add_lit.tex
{\centering
\definecolor{addlitgeneral}{HTML}{008300}%
\definecolor{addlitselfimprove}{HTML}{4A3AA7}%
\definecolor{addplacebo}{HTML}{8F8E89}%
\definecolor{addnone}{HTML}{52514E}%
\hspace{0pt}\lgdentry{addlitgeneral}{0.90pt}{}{lit-general}\hfill \lgdentry{addlitselfimprove}{0.90pt}{}{lit-selfimprove}\hfill \lgdentry{addplacebo}{0.90pt}{dash pattern=on 0.62pt off 1.02pt}{placebo}\hfill \lgdentry{addnone}{0.90pt}{dash pattern=on 2.30pt off 0.99pt}{none}\hspace{0pt}
\par}

%% file: figures/tables/table_add_lit_2048.tex
\begin{tabular}{lrrr}
\toprule
Values & Held-out & $\delta$ & $p$ \\
\midrule
lit-general & 119,565 & $-$0.22 & 0.393 \\
lit-selfimprove & 107,936 & \textbf{$-$0.50} & 0.055 \\
placebo & 103,561 & \textbf{$-$0.50} & \textbf{0.037} \\
none & 126,177 & -- & -- \\
\bottomrule
\end{tabular}

%% file: figures/legends/legend_add_lit_2048.tex
{\centering
\definecolor{addlitgeneral}{HTML}{008300}%
\definecolor{addlitselfimprove}{HTML}{4A3AA7}%
\definecolor{addplacebo}{HTML}{8F8E89}%
\definecolor{addnone}{HTML}{52514E}%
\hspace{0pt}\lgdentry{addlitgeneral}{0.90pt}{}{lit-general}\hfill \lgdentry{addlitselfimprove}{0.90pt}{}{lit-selfimprove}\hfill \lgdentry{addplacebo}{0.90pt}{dash pattern=on 0.62pt off 1.02pt}{placebo}\hfill \lgdentry{addnone}{0.90pt}{dash pattern=on 2.30pt off 0.99pt}{none}\hspace{0pt}
\par}

%% file: figures/tables/table_add_disp.tex
\begin{tabular}{lrrr}
\toprule
Values & Held-out & $\delta$ & $p$ \\
\midrule
openness & 571,927 & -- & -- \\
openness-deon & 559,659 & \textbf{$-$0.54} & \textbf{0.024} \\
openness-util & 561,063 & \textbf{$-$0.50} & \textbf{0.023} \\
openness-virt & 566,676 & $-$0.30 & 0.095 \\
conservation & 511,544 & -- & -- \\
conservation-deon & 478,644 & \textbf{$-$0.52} & 0.100 \\
conservation-util & 524,396 & +0.40 & 0.422 \\
conservation-virt & 519,751 & +0.08 & 0.541 \\
\bottomrule
\end{tabular}

%% file: figures/legends/legend_add_disp.tex
{\centering
\definecolor{addopenness}{HTML}{179568}%
\definecolor{addopennessdeon}{HTML}{4AAC89}%
\definecolor{addopennessutil}{HTML}{7DC4AA}%
\definecolor{addopennessvirt}{HTML}{B0DBCC}%
\definecolor{addconservation}{HTML}{2466B6}%
\definecolor{addconservationdeon}{HTML}{5488C6}%
\definecolor{addconservationutil}{HTML}{84A9D6}%
\definecolor{addconservationvirt}{HTML}{B5CBE6}%
\hspace{0pt}\lgdentry{addopenness}{0.90pt}{}{openness}\hfill \lgdentry{addopennessdeon}{0.90pt}{dash pattern=on 2.30pt off 0.99pt on 0.62pt off 0.99pt}{openness-deon}\hfill \lgdentry{addopennessutil}{0.90pt}{dash pattern=on 2.30pt off 0.99pt}{openness-util}\hfill \lgdentry{addopennessvirt}{0.90pt}{dash pattern=on 0.62pt off 1.02pt}{openness-virt}\hspace{0pt}\\[2pt]
\hspace{0pt}\lgdentry{addconservation}{0.90pt}{}{conservation}\hfill \lgdentry{addconservationdeon}{0.90pt}{dash pattern=on 2.30pt off 0.99pt on 0.62pt off 0.99pt}{conservation-deon}\hfill \lgdentry{addconservationutil}{0.90pt}{dash pattern=on 2.30pt off 0.99pt}{conservation-util}\hfill \lgdentry{addconservationvirt}{0.90pt}{dash pattern=on 0.62pt off 1.02pt}{conservation-virt}\hspace{0pt}
\par}

%% file: figures/tables/table_add_disp_texts.tex
\begin{tabular}{@{}lrlrl@{}}
\toprule
 & \multicolumn{2}{c}{openness} & \multicolumn{2}{c}{conservation} \\
\cmidrule(lr){2-3}\cmidrule(lr){4-5}
Framing & Words & Excerpt & Words & Excerpt \\
\midrule
\begin{tabular}[t]{@{}l@{}}original\\\textit{Priorities}\end{tabular} & 133 & \parbox[t]{137pt}{\raggedright prefer the bold transformation \ldots{} over the safe incremental tweak} & 137 & \parbox[t]{137pt}{\raggedright reuse the proven approach over a novel one} \\
\addlinespace[3pt]
\begin{tabular}[t]{@{}l@{}}deontological\\\textit{Rules}\end{tabular} & 175 & \parbox[t]{137pt}{\raggedright You MUST take the bold transformation \ldots{} over the safe incremental tweak.} & 174 & \parbox[t]{137pt}{\raggedright You MUST reuse the proven approach over a novel one.} \\
\addlinespace[3pt]
\begin{tabular}[t]{@{}l@{}}utilitarian\\\textit{reward}\end{tabular} & 173 & \parbox[t]{137pt}{\raggedright A bold transformation \ldots{} outweighs the safe incremental tweak} & 169 & \parbox[t]{137pt}{\raggedright reusing a proven approach outweighs a novel one} \\
\addlinespace[3pt]
\begin{tabular}[t]{@{}l@{}}virtue\\\textit{Character}\end{tabular} & 175 & \parbox[t]{137pt}{\raggedright you are restless \ldots{} You enjoy taking apart what you judge settled too soon.} & 176 & \parbox[t]{137pt}{\raggedright you are careful \ldots{} You enjoy making safe what you judge loose.} \\
\bottomrule
\end{tabular}

%% file: figures/tables/table_add_model.tex
\begin{tabular}{lrrr}
\toprule
Values & Held-out & $\delta$ & $p$ \\
\midrule
openness-opus & 571,927 & -- & -- \\
openness-haiku & 292,861 & \textbf{$-$1.00} & \textbf{$<$0.001} \\
openness-sonnet & 478,869 & \textbf{$-$1.00} & \textbf{$<$0.001} \\
conservation-opus & 511,544 & -- & -- \\
conservation-haiku & 296,058 & \textbf{$-$1.00} & \textbf{$<$0.001} \\
conservation-sonnet & 393,062 & \textbf{$-$0.96} & \textbf{$<$0.001} \\
\bottomrule
\end{tabular}

%% file: figures/legends/legend_add_model.tex
{\centering
\definecolor{addopenness}{HTML}{1BAF7A}%
\definecolor{addopennesshaiku}{HTML}{1BAF7A}%
\definecolor{addopennesssonnet}{HTML}{1BAF7A}%
\definecolor{addconservation}{HTML}{2A78D6}%
\definecolor{addconservationhaiku}{HTML}{2A78D6}%
\definecolor{addconservationsonnet}{HTML}{2A78D6}%
\hspace{0pt}\lgdentry{addopenness}{0.90pt}{}{openness-opus}\hfill \lgdentry{addopennesshaiku}{0.90pt}{dash pattern=on 2.30pt off 0.99pt}{openness-haiku}\hfill \lgdentry{addopennesssonnet}{0.90pt}{dash pattern=on 0.62pt off 1.02pt}{openness-sonnet}\hspace{0pt}\\[2pt]
\hspace{0pt}\lgdentry{addconservation}{0.90pt}{}{conservation-opus}\hfill \lgdentry{addconservationhaiku}{0.90pt}{dash pattern=on 2.30pt off 0.99pt}{conservation-haiku}\hfill \lgdentry{addconservationsonnet}{0.90pt}{dash pattern=on 0.62pt off 1.02pt}{conservation-sonnet}\hspace{0pt}
\par}